\pdfoutput=1
\documentclass{article}
\usepackage[final]{corl_2020} % Uncomment for the camera-ready ``final'' version.
\usepackage[numbers]{natbib}
\usepackage{authblk}
\usepackage{multicol}
\usepackage{mathtools}
\usepackage{dblfloatfix}
\usepackage{amsmath}
\usepackage{amsfonts}
\usepackage{caption}
\usepackage{wrapfig}
\usepackage{enumitem}
\usepackage{amsmath}
\usepackage{longtable}
\usepackage{subcaption}
\usepackage[para]{footmisc}
\usepackage{amssymb}
\usepackage[dvipsnames]{xcolor}

\renewcommand{\vec}[1]{\boldsymbol{#1}}

\usepackage{tikz}
\usetikzlibrary{bayesnet}

\usepackage{color}
\usepackage{graphicx}
\usepackage{caption}
\usepackage{subcaption}
\usetikzlibrary{backgrounds}
\usetikzlibrary{arrows.meta}
\usepackage{lipsum}

\newcommand\blfootnote[1]{%
  \begingroup
  \renewcommand\thefootnote{}\footnote{#1}%
  \addtocounter{footnote}{-1}%
  \endgroup
}

\newcommand{\tikzACRKN}{
\begin{tikzpicture}[thick]
		\tikzset{invisible/.style={draw=none},
	}

    \draw[orange] (0.25, 2) -- (1.75, 2) -- (2.25, 3.0) -- (-0.25, 3.0) -- (0.25, 2.0)  node[black, pos=.5, xshift=27.5]{Encoder};
    \draw[->, >=stealth] (1.0, 2.0) -- (1.0, 0.85) node[pos=.35, xshift=5]{$\vec{w}_t ~~ \vec{\sigma}^\textrm{obs}_t$};
    \draw[->, >=stealth] (1.0, 3.5) -- (1.0, 3.0) node[pos=.2, xshift=5]{$~~\vec{o}_t$};
    
    \draw[->, >=stealth] (4.5, 2.5) -- (4.5, 0.85) node[pos=.2, xshift=-10, align=left]{$\vec{a}_t$};
    
	\draw[densely dotted, black!60!green] (-0.25, -0.4)   rectangle  (5.75, 1.3) node[black, pos=0.1, xshift=10]{RKN Cell};
	\draw (3.5, 0.15) rectangle (5.5, 0.85) node[pos=.5, align=center]{Predict};
	\draw (0.0, 0.15) rectangle (2.0, 0.85) node[pos=.5, align=center]{Update}; 
    \draw[->, >=stealth] (-1.5, 0.5) -- (0.0, 0.5) node[pos=.25, yshift=11]{$\vec{z}_t^-$, $\vec{\Sigma}_t^-$};
    \draw[->, >=stealth] (-1.5, 0.5) -- (0.0, 0.5) node[pos=.25, yshift=-8]{\small{Latent State}};
    \draw[->, >=stealth] (2.0, 0.5) -- (3.5, 0.5) node[pos=.5, yshift=11]{$\vec{z}_t^+$, $\vec{\Sigma}_t^+$};
    \draw[->, >=stealth] (5.5, 0.5) -- (7.0, 0.5) node[pos=.75, yshift=11]{$\vec{z}_{t+1}^-$, $\vec{\Sigma}_{t+1}^-$};

    \draw[->, >=stealth] (6.5, 0.5) -- (6.5, -1.0); 
    
    \draw[blue] (5.75, -1.0) -- (7.25, -1.0) -- (7.75, -2.0) -- (5.25, -2.0) -- (5.75, -1.0) node[black, pos=.5, xshift=27.5]{Decoder};

    \draw[->, >=stealth] (6.5, -2.0) -- (6.5, -2.5) node[pos=0.6, xshift=13]{$\hat{\vec{o}}_{t+1}$}; 
\end{tikzpicture}
}

\newcommand{\tikzACRKNINVDYN}{
\begin{tikzpicture}[thick]
		\tikzset{invisible/.style={draw=none},
	}

    \draw[orange] (0.25, 2) -- (1.75, 2) -- (2.25, 3.0) -- (-0.25, 3.0) -- (0.25, 2.0)  node[black, pos=.5, xshift=27.5]{Encoder};
    \draw[->, >=stealth] (1.0, 2.0) -- (1.0, 0.85) node[pos=.35, xshift=5]{$\vec{w}_t ~~ \vec{\sigma}^\textrm{obs}_t$};
    \draw[->, >=stealth] (1.0, 3.5) -- (1.0, 3.0) node[pos=.2, xshift=5]{$~~\vec{o}_t$};
    
    \draw[->, >=stealth] (4.5, 2.5) -- (4.5, 0.85) node[pos=.2, xshift=25, align=left, black!30!red]{\small{Executed}\\\small{Action}};
    \draw[->, >=stealth] (4.5, 2.5) -- (4.5, 0.85) node[pos=.2, xshift=-10, align=left]{$\vec{a}_t$};
    
	\draw[densely dotted, black!60!green] (-0.25, -0.4)   rectangle  (5.75, 1.3) node[black, pos=0.1, xshift=10]{RKN Cell};
	\draw (3.5, 0.15) rectangle (5.5, 0.85) node[pos=.5, align=center]{Predict};
	\draw (0.0, 0.15) rectangle (2.0, 0.85) node[pos=.5, align=center]{Update}; 
    \draw[->, >=stealth] (-1.5, 0.5) -- (0.0, 0.5) node[pos=.25, yshift=11]{$\vec{z}_t^-$, $\vec{\Sigma}_t^-$};
    \draw[->, >=stealth] (-1.5, 0.5) -- (0.0, 0.5) node[pos=.25, yshift=-8]{\small{Latent State}};
    \draw[->, >=stealth] (2.0, 0.5) -- (3.5, 0.5) node[pos=.5, yshift=11]{$\vec{z}_t^+$, $\vec{\Sigma}_t^+$};
    \draw[->, >=stealth] (5.5, 0.5) -- (7.0, 0.5) node[pos=.75, yshift=11]{$\vec{z}_{t+1}^-$, $\vec{\Sigma}_{t+1}^-$};

    \draw[->, >=stealth] (2.75, 0.5) -- (2.75, -1.0);
    \draw[->, >=stealth] (6.5, 0.5) -- (6.5, -1.0); 
    
    \draw[blue] (5.75, -1.0) -- (7.25, -1.0) -- (7.75, -2.0) -- (5.25, -2.0) -- (5.75, -1.0) node[black, pos=.5, xshift=27.5]{Decoder};
    \draw[black!30!red] (2.0, -1.0) -- (3.5, -1.0) -- (4.0, -2.0) -- (1.5, -2.0) -- (2.0, -1.0) node[black, pos=.5, xshift=27.5, align=center]{Action\\Decoder};
    
    \draw[->, >=stealth] (2.75, -2.) -- (2.75, -2.5) node[pos=0.6, xshift=10]{$\hat{\vec{a}}_t$};
    
    \draw[->, >=stealth] (0, -1.5) -- (1.75, -1.5) node[black!30!red, pos=0.0, align=center]{\small{(Desired) Next}\\ \small{Observation}};
    
    \draw[->, >=stealth] (6.5, -2.0) -- (6.5, -2.5) node[pos=0.6, xshift=13]{$\hat{\vec{o}}_{t+1}$}; 
\end{tikzpicture}
}

 \tikzset{%
    dashedarrow/.style = {%
        draw,  > = {Latex[width = 2.5mm, length = 3.5mm, open, fill = white]}, ->
    }%
}% 
\newcommand{\tikzRKNGM}{ \begin{tikzpicture}[very thick,scale=1.0, every node/.style={scale=1.0}]
    \node[latent, fill, minimum size=42pt] (s1) {\LARGE$\vec{z}_t$}; %
    \node[latent,right=of s1, minimum size=42pt, xshift=2cm] (s2) {\LARGE$\vec{z}_{t+1}$};
    \coordinate[left of=s1,xshift=-1cm] (c1);
    \coordinate[right of=s2,xshift=1cm] (c2);

     \node[latent,below=of s1, fill,minimum size=42pt] (h1) {\LARGE$\vec{w}_t$}; 
     \node[latent,below=of s2, fill,minimum size=42pt] (h2) {\LARGE$\vec{w}_{t+1}$};

     \node[obs,below=of h1, xshift=-1cm,yshift=0cm,minimum size=42pt] (o1) {\LARGE$\vec{o}_{t}$};
     \node[obs,below=of h2,xshift=-1cm,yshift=0cm,minimum size=42pt] (o2) {\LARGE$\vec{o}_{t+1}$};
    
     \node[obs,below=of h1, minimum size=42pt,xshift=1cm
     ] (a1) {\LARGE$\vec{a}_{t}$};%
     \node[obs,below=of h2, minimum size=42pt,xshift=1cm
     ] (a2) {\LARGE$\vec{a}_{t+1}$};%

     \edge{s1} {h1}
     \path[dashedarrow] (o1) edge[] (h1);
    \path[dashedarrow] (a1) edge[] (h1);
     \edge {s1} {s2}
     \edge {s2} {h2}
    \path[dashedarrow] (o2) edge[] (h2);
    \path[dashedarrow] (a2) edge[] (h2);
     \edge[dashed] {s2} {c2}
     \edge[dashed] {c1} {s1}
     \end{tikzpicture}
}

\newcommand{\tikzACRKNGM}{ \begin{tikzpicture}[very thick,scale=1.0, every node/.style={scale=1.0}]
    \node[latent, fill, minimum size=42pt] (s1) {\LARGE$\vec{z}_t$}; %
    \node[latent,right=of s1, minimum size=42pt, xshift=2cm] (s2) {\LARGE$\vec{z}_{t+1}$};
    \coordinate[left of=s1,xshift=-1cm] (c1);
    \coordinate[right of=s2,xshift=1cm] (c2);

     \node[latent,below=of s1, fill,minimum size=42pt] (h1) {\LARGE$\vec{w}_t$}; 
     \node[latent,below=of s2, fill,minimum size=42pt] (h2) {\LARGE$\vec{w}_{t+1}$};

     \node[obs,below=of h1, minimum size=42pt] (o1) {\LARGE$\vec{o}_{t}$};
     \node[obs,below=of h2, minimum size=42pt] (o2) {\LARGE$\vec{o}_{t+1}$};
    
    \coordinate[right of=s2,xshift=1cm,yshift=1.0cm] (c3);
    
     \node[obs,above=of s1,yshift=-0.3cm,minimum size=42pt] (a1) {\LARGE $\vec{a}_t$};
     \node[obs,above=of s2,yshift=-0.3cm,minimum size=42pt] (a2) {\LARGE $\vec{a}_{t+1}$};
    
     \edge {s1} {h1}
     \path[dashedarrow] (o1)edge[] (h1);
     \edge {a1} {s2}
     \edge {s1} {s2}
     \edge {s2} {h2}
     \path[dashedarrow] (o2) edge[] (h2);
     \edge[dashed] {s2} {c2}
     \edge[dashed] {a2} {c3}
 \edge[dashed] {c1} {s1}
 \end{tikzpicture}
 }

\title{Action-Conditional Recurrent Kalman Networks For Forward and Inverse Dynamics Learning}

\author[1,2]{\textbf{Vaisakh Shaj}}
\author[1]{\textbf{Philipp Becker}}
\author[3]{\textbf{Dieter Büchler}}
\author[2]{\textbf{Harit Pandya}}
\author[4]{\textbf{Niels van Duijkeren}}
\author[5]{\mbox{\textbf{C. James Taylor}}}
\author[2]{\textbf{Marc Hanheide}}
\author[1]{\textbf{Gerhard Neumann}}
\affil[1]{Autonomous Learning Robots, KIT, Germany}
\affil[2]{LCAS, University Of Lincoln, UK}
\affil[3]{Max Planck Institute for Intelligent Systems, Tübingen, Germany}
\affil[4]{Bosch Corporate Research, Renningen, Germany}
\affil[5]{Lancaster University, UK}
\begin{document}
\maketitle

%===============================================================================

\begin{abstract}
Estimating accurate forward and inverse dynamics models is a crucial component of model-based control for sophisticated robots such as robots driven by hydraulics, artificial muscles, or robots dealing with different contact situations.
Analytic models to such processes are often unavailable or inaccurate due to complex hysteresis effects, unmodelled friction and stiction phenomena, and unknown effects during contact situations.
A promising approach is to obtain spatio-temporal models in a data-driven way using recurrent neural networks, as they can overcome those issues.
However, such models often do not meet accuracy demands sufficiently, degenerate in performance for the required high sampling frequencies and cannot provide uncertainty estimates.  

We adopt a recent probabilistic recurrent neural network architecture, called Recurrent Kalman Networks (RKNs), to model learning by conditioning its transition dynamics on the control actions.
RKNs outperform standard recurrent networks such as LSTMs on many state estimation tasks.
Inspired by Kalman filters, the RKN provides an elegant way to achieve action conditioning within its recurrent cell by leveraging additive interactions between the current latent state and the action variables.
We present two architectures, one for forward model learning and one for inverse model learning.
Both architectures significantly outperform existing model learning frameworks as well as analytical models in terms of prediction performance on a variety of real robot dynamics models.
\end{abstract}


%===============================================================================

\section{Introduction}
\blfootnote{Correspondence to Vaisakh Shaj $<$v.shaj@kit.edu$>$}
Dynamics models are an integral part of many control architectures.
Depending on the control approach, the control law relies either on a forward model, mapping from control input to the change of system state, or on an inverse model, mapping from the change of the system state to control signals.
However, analytical dynamics models are either not available or often too inaccurate in situations such as robots driven by hydraulics, artificial muscles, or robots dealing with unknown contact situations. 
Hence, there is a significant need for data-driven approaches that can deal with such complex systems. 
Yet, modelling dynamics is a challenging task due to the inherent hysteresis effects, unmodelled friction and stiction phenomena, and unknown properties of the interacting objects. 
Additional challenges for modelling are given by the high data frequency, often up to 1kHz. 
Furthermore, many modern model-based architectures~\cite{deisenroth2011pilco}\cite{chua2018deep} for learning controllers rely on uncertainty estimates of the prediction. 
Hence, such probabilistic modelling ability is another point on the desiderata for model learning algorithms.  

In this paper, we extend a recent recurrent neural network architecture~\cite{becker2019recurrent}, called Recurrent Kalman Networks (RKN), for learning forward and inverse dynamics models.
The RKN provides probabilistic predictions of future states, can deal with noisy, high dimensional and missing inputs, and has been shown to outperform Long-Short Term-Memory Networks (LSTMs)\cite{lstm} and Gated Recurrent Units (GRUs) \cite{gru} on many state estimation tasks. The authors~\cite{becker2019recurrent} focussed on estimating the state of the system from images.
%We adapt the RKN architecture to be used for prediction instead of filtering and show that it significantly outperforms state-of-the-art models for learning robot forward and inverse dynamics, where the observations are typically low dimensional, i.e., positions and velocities of the joints of the robot, but occur with a much higher frequency.
We adapt the RKN architecture for prediction instead of filtering and employ it for learning robot forward and inverse dynamics. 
In this scenario, the observations are typically low dimensional, i.e., positions and velocities of the joints of the robot, but occur with a much higher frequency.
Unlike original RKN, we explicitly model the control actions in our architecture.  
%This action-conditioning is achieved by modifying the transition dynamics in the latent space using an action dependent additive factor for learning action conditioned models.
We modify the model of the latent transition dynamics using an action dependent non-linear additive factor in order to condition them with action variables.
We call this architecture action-conditional RKN (ac-RKN) and show that our approach is more accurate than competing learning methods and even analytical models where they are available. 
Moreover, we extend the ac-RKN to learning inverse dynamics by adding an action decoder network.
Also our learned inverse dynamics models outperform analytical models and competitive learning methods.

%===============================================================================

\section{Related Works}
\label{sec:citations}

\textbf{Robot Dynamics Learning}. %Comment Geri: Thats already part of the intro, we do not need it twice. Models describing system dynamics, i.e., the coupling of control input $a$ and system state $s$, are essential for model-based control approaches~\cite{ioannou2012robust}. Depending on the control approach, the control law relies either on the forward model $f$, mapping from current state and control input to the next state, or on the inverse model $f^{-1}$, mapping from system change to control input. 
Due to the increasing complexity of robot systems, analytical models are more difficult to obtain. This problem leads to a variety of model estimation techniques which allow the roboticist to acquire models from data.
When learning the model, mostly standard regression techniques with Markov assumptions on the states and actions are applied to fit either the forward or inverse dynamics model to the training data.
Authors used Linear Regression~\cite{schaal2002scalable}\cite{haruno2001mosaic}, Gaussian Mixture Regression~\cite{calinon2010learning}\cite{khansari2011learning}, Gaussian Process Regression~\cite{nguyen2009model}\cite{nguyen2010using}, Support Vector Regression~\cite{ferreira2007simulation}, and feed forward neural networks~\cite{polydoros2015real}. 
However the Markov assumptions are violated in many cases, e.g., for learning the dynamics of complex robots like hydraulically and pneumatically actuated robots or soft robots.
Here, the Markov assumptions for states and actions no longer hold due to hysteresis effects and unobserved dependencies. 
Similarly, learning the dynamics of robots in frictional contacts with unknown objects also violates the Markov assumption and requires reliance on recurrent models which can take into account the temporal dynamic behavior of input sequences while making predictions. 
Recently, using RNNs such as LSTMs or GRUs has been attempted. 
Those scale easily to big data, available due to the high data frequencies, and allow learning in $O(n)$ time~\cite{rueckert2017learning}. 
However, the lack of a principled probabilistic modelling and action conditioning makes them less reliable to be applied for robotic control applications. 

\textbf{Action-Conditional Probabilistic Models}. In model based control and reinforcement learning (RL) problems, learning to predict future states and observations conditioned on actions is a key component.
Approaches such as~\cite{deisenroth2011pilco}, \cite{nagabandi2018neural}, \cite{lenz2015deepmpc}, \cite{oh2015action}, \cite{finn2016unsupervised} try to achieve this by using traditional models like GPs, feed forward neural networks, or LSTMs. 
The action conditioning is realized by concatenating the input observations and action signals~\cite{nagabandi2018neural}\cite{deisenroth2011pilco} or via factored conditional units~\cite{taylor2009factored}, where features from some number of inputs modulate multiplicatively and are then weighted to form network outputs. 
In each of these approaches action conditioning happens outside of the recurrent neural network cell which leads to sub-optimal performance as observations and actions are treated similarly. 
We propose an action-conditional RNN cell inspired by Kalman filter operations~\cite{becker2019recurrent} which provide an elegant way to learn not just forward but regularized inverse models as well, giving significantly better performance compared to the current state of the art.
Also, opposed to previous approaches for dynamics learning we evaluate not only on relatively simple electrically actuated robotic systems but on a wide variety of complex robotic systems, including large hydraulically actuated robots and pneumatically actuated soft robots for one-step-ahead and long term prediction tasks.
%===============================================================================

\section{Recurrent Kalman Networks}
Recurrent Kalman Networks (RKN) \cite{becker2019recurrent} integrate uncertainty estimates into deep time-series modelling by combining Kalman Filters \cite{kalman1960new} with deep neural networks. 
To efficiently address highly nonlinear dynamics the RKN employs a Kalman filter in a high dimensional, factorized latent space where the dynamics can be modelled using locally linear models.
This space is learned by an encoder-decoder structure mapping from and back to the latent representation. 
Both encoder and decoder is in practice realized as deep neural networks.
However, the encoder does not only map the observation to the latent space but also emits an estimate of the uncertainty in the observation. 
This uncertainty estimate is used, together with latent state uncertainties inherently present in the Kalman filter, to control the update of the state/memory of the recurrent cell. 
This update procedure for the memory is in contrast to traditional deep recurrent approaches, such as LSTMs \cite{lstm}, where input and forget gates have a similar purpose but do not offer a probabilistic interpretation. 
Formally, the RKN first uses a locally linear dynamics model, $\vec{A}_{t-1}$, to advance the last posterior belief $\mathcal{N}(\vec{z}_{t-1}^+, \vec{\Sigma}^{+}_{t-1})$ in time and obtain a prior belief $\mathcal{N}(\vec{z}_t^-, \vec{\Sigma}^-_t)$ for the current time step $t$. 
Here, $\vec{A}_{t-1}$ depends on the posteriors mean, $\vec{z}_{t-1}^+$, i.e, $\vec{A}_{t-1} := \vec{A}(\vec{z}^+_{t-1})$, and the transition noise is modelled by $\mathcal{N}(\vec{0}, \vec{I} \cdot \vec{\sigma^\textrm{trans}})$, with a learned, vector value $\vec{\sigma^\textrm{trans}}$.
This prior belief is then updated using a probabilistic representation of the observation $\mathcal{N}(\vec{w}_t, \vec{\sigma}^\textrm{obs}_t)$, provided by the encoder. Classical Kalman update equations then infer the Kalman gain $\vec{Q}_t$ and, subsequently, the current posterior $\mathcal{N}(\vec{z}_t^+, \vec{\Sigma}^{+}_t)$. Ultimately, the decoder maps from the posterior to the desired output, where we can learn separate decoders for the mean and the variance estimates. The network is trained end to end via backpropagation through time.    

A detailed description of the RKN, the exact factorization assumptions and parameterization of the locally linear transition model is given in \cite{becker2019recurrent} and is also summarized in the appendix.

\section{Robot Dynamics Learning with Action-Conditional RKNs}

%\subsection{Action Conditional Reccurent Kalman Networks}
In this paper, we are interested in learning accurate forward or inverse dynamics models of complex robots with non-Markovian dynamics. 
Learning such models requires action-dependent prediction models of high accuracy in order to be useful for robot control.
Moreover, we deal with systems with high sample frequencies, up to $1$kHz. 
We extend the RKN approach to robot dynamics learning with the following innovations:
\textbf{(i)} We use an \textbf{action-conditional prediction} update which provides a natural way to incorporate control inputs into the latent dynamical system. %The action conditioning is implemented as additive interactions of the control signal and the previous latent state.
%\item As opposed to \cite{becker2019recurrent}, we concentrate on prediction instead of filtering. To do so, the RKN architecture needs to be changed slightly such that we decode the prior belief state instead of the posterior belief state.
% \item \textbf{Difference Predictor.} We predict the state differences instead of the absolute state values, which is crucial in particular for high sample frequencies. 
%We therefore resort to predicting state differences and introduce an additional memory architecture that can integrate predicted state differences over time in case of missing observations. 
%item \textbf{Forward Dynamics.} 
\textbf{(ii)} We modify the architecture to perform prediction instead of filtering and learn \textbf{accurate forward dynamics models} for multiple robots with complex actuator dynamics. 
%item \textbf{Inverse Dynamics.} 
\textbf{(iii)} We incorporate an inverse dynamics decoder in the RKN architecture, allowing us to learn jointly forward and \textbf{inverse dynamics models}.
%\end{enumerate}
We refer to the resulting approach as action-conditional RKN (ac-RKN).
%We call our model action conditional RKN (ac-RKN). % due to the action conditioned prediction step.
%The architecture for the ac-RKN for learning forward dynamics is presented in Figure 1. 

%As seen in the figure, an encoder network extracts latent features $\vec{w}_t$ from the current observation $\vec{o}_t$. Additionally, it emits an estimate of the uncertainty in the features via the variance vector $\vec{\sigma}_t^\mathrm{obs}$.\\

\subsection{Action Conditioning}
\label{sec:action-con}
%Both quantities are used to to update the latent prior from the previous time step $\left(\vec{z}_{t}^-, \vec{\Sigma}_{t}^- \right)$ to the posterior estimate $\left(\vec{z}_{t}^+, \vec{\Sigma}_{t}^+ \right)$. This operation is conceptually similar to an input gate in an LSTM, which decides what new information would be stored in the latent state based on the current observation. However unlike LSTMs the gating is governed by uncertainties. Intuitively, the higher the uncertainty in the current observation (e.g., due to noise or missing data), the lesser would be its contribution to the latent state. As discussed in section A, 

To achieve action conditioning within the recurrent cell, we modify the transition model proposed in \cite{becker2019recurrent} to include  a control model $\vec{b}(\vec a_t)$ in addition to the locally linear transition model $\vec{A}_t$. 
The control model $\vec{b}(\vec a_t)$ is added to the locally
linear state transition, i.e., $\vec{z}^-_{t+1} = \vec{A}_t \vec{z}^+_{t} + \vec b(\vec a_t)$. %where $\vec{z}^+_{t}$ is the posterior mean at time step $t$ and $\vec{z}^-_{t+1} $ the prior mean for time step $t+1$. 
As illustrated in \autoref{fig:pgm}, this formulation of latent dynamics captures the causal effect of the actions variables $\vec{a}_t$ on the state transitions in a more principled manner than including the action as an observation, which is the common approach for action conditioning in RNNs \cite{becker2019recurrent}. 
The control model $\vec{b}(\vec a_t)$ can be represented in several ways, i.e.:
\begin{wrapfigure}[14]{r}{.55\linewidth}
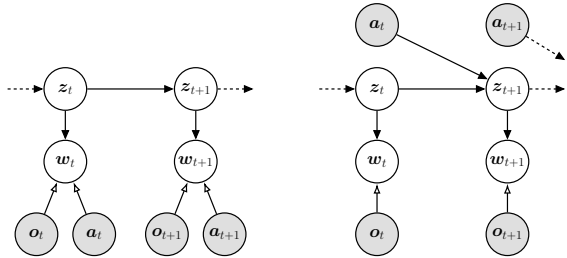

%      \hspace*{0.75cm} 
%      \includegraphics[scale=0.5]{Figures/RKN-SchematicArrow.pdf}
 %     \caption{Schematic Diagram Of Action Conditional Recurrent Kalman Network. Action conditioning takes place via additive interactions between the latent state $\vec{z}_{t}$ and action $a_t$ at the predict stage in the RKN Cell.}

%     \hspace*{0.2cm} 
   %   \includegraphics[scale=0.4]{Figures/ForwardArrow.pdf}
%   \begin{subfigure}[b]{0.45\textwidth}
%     \centering
%      \resizebox{\textwidth}{!}{\tikzACRKN}
%      \end{subfigure}
% \qquad     
    \begin{subfigure}[b]{0.24\textwidth}
    \centering
     \resizebox{\textwidth}{!}{\tikzRKNGM}
     %\caption{}
     \end{subfigure}
\qquad
         \begin{subfigure}[b]{0.24\textwidth}
     \centering
     \resizebox{\textwidth}{!}{\tikzACRKNGM}
     %\caption{}
     \end{subfigure}
% \qquad
% \begin{subfigure}[b]{0.5\textwidth}
%      \centering
%      \resizebox{\textwidth}{!}{\tikzACRKN}
%          \caption{}
%      \end{subfigure}
\caption{ Graphical models for actions treated as fake observations as in \cite{becker2019recurrent} (left) and ac-RKN (right) which treats actions in a principled manner by capturing its causal effect on the state transition via additive interactions with the latent state. The hollow arrows denote deterministic transformation leading to implicit distributions.} 
 \label{fig:pgm}
 \end{wrapfigure}
\begin{itemize}[leftmargin=0.5cm]
\item[\textbf{(i)}] \textbf{Linear:} $\vec{b}_{l}(\vec a_t) =\vec{B} \vec a_t$, where $\vec{B} $ is a linear transformation matrix.
\item[\textbf{(ii)}] \textbf{Locally-Linear:} $\vec{b}_{m}(\vec a_t) = \vec{B}_t \vec a_t$, where $\vec{B}_t = \sum_{k=0}^K \beta^{(k)}(\vec{z_t}) \vec{B}^{(k)}$ is a linear combination of k linear control models $\vec{B}^{(k)}$. A  small  neural  network  with softmax  output  is  used  to  learn $\beta^{(k)}$.  
\item[\textbf{(iii)}] \textbf{Non-Linear:} $\vec{b}_n(\vec a_t) = \vec{f}(\vec a_t)$, where $\vec{f}(.)$ can be any non-linear function approximator. We use a multi-layer neural network regressor with ReLU activations.
\end{itemize} 

Note that the prediction step for the variance is not affected by this choice of action conditioning, i.e., $\vec{\Sigma}^-_{t+1} = \vec{A}_t \vec{\Sigma}_{t}^+ \vec{A}_t^T + \vec{I} \cdot \vec{\sigma^\mathrm{trans}}$ as the action is known and not uncertain.
Thus, unlike the state transition model $\vec{A}_t$, we do not need to constrain the model $ \vec{b}$ to be linear or locally-linear as it neither affects the Kalman gain nor how the covariances are updated. 
%Hence, ac-RKN uses $\vec{b}_n(\vec a_t)$ as the control model, which provides a more flexible non-linear transformation and showed the best performance as shown in our experiments. 
Hence, we use the non-linear approach, $\vec{b}_n(\vec a_t)$, as it provides the most flexibility and achieved the best performance as shown in section \ref{sec:fwd_dyn_res}.
The Kalman update step is also unaffected by the new action-conditioned prediction step and therefore remains as presented in~\cite{becker2019recurrent}.

%described in the previous section.  %Comment: Experiments should come later in the exp. section. Figure \ref{fig:control} shows the performance of each of these control models for multi-step ahead prediction tasks.

Our principled treatment of control signals is crucial for learning accurate forward dynamics models, as seen in section \ref{sec:forwardDyn}. Moreover, the disentangled representation of actions in the latent space gives us more flexibility in manipulating the control actions for different applications including inverse dynamics learning (section \ref{sec:inverseDyn}), extensions to model-based reinforcement learning and planning. For long-term prediction using different action sequences, we can still apply the latent transition dynamics without using observations for future time steps.

%\textbf{Action Conditional Prediction Equations}.
%The prior for the next time step $\left(\vec{z}^-_{t+1},  \vec{\Sigma}^-_{t+1} \right)$ is obtained from the current posterior $\left(\vec{z}_{t}^+, \vec{\Sigma}_{t}^+ \right)$ by
%\begin{equation}\begin{multlined}\vec{z}^-_{t+1} = \vec{A}_t \vec{z}^+_{t} + b(a_t) &  \vec{\Sigma}^-_{t+1} = \vec{A}_t \vec{\Sigma}_{t}^+ \vec{A}_t^T + \vec{I} \cdot \vec{\sigma}^\mathrm{trans}
%\end{multlined}
%\end{equation}
%, where $b$ is a feed forward neural network function approximator with relu activation in hidden layer and linear activation in the output layer. The locally linear transition model $\vec{A}_t$ has a similar structure as discussed in Section \RomanNumeralCaps{3} A.

%The current latent state $\vec{z}_t$ consists of the observable units $\vec{p}_t$ as well as the corresponding memory units $\vec{m}_t$.
%Finally, a decoder produces either $\left(\vec{\hat{o}}_{t+1}, \vec{\hat{\sigma}}_{t+1} \right)$, a low dimensional observation and an element-wise uncertainty estimate.

%In conclusion we extend the RKN in \cite{becker2019recurrent} to enable action conditioning in a principled manner by an additive interaction of the action/control at the predict stage of the transition cell in RKN. 
%Also since we use this for forward dynamics learning which is basically a prediction task rather than filtering we decode the prior which gave slightly better performance than the architecture present in \cite{becker2019recurrent} 
%where the posterior is decoded to the desired output space.

\subsection{Forward Dynamics Learning}
\label{sec:forwardDyn}

We want to learn a forward dynamics model $f: \vec o_{1:t}, \vec  a_{1:t} \mapsto \vec o_{t+1}$ that predicts the next observation $\vec o_{t+1}$ given histories of observation $\vec o_{1:t}$ and actions $\vec a_{1:t}$. 
We explicitly focus on non-Markovian systems, where the observations are typically joint angles and, if available, joint velocities of all degrees of freedom of the robot. 
In our experiments, we aim to predict the joint angles at the next time step given a sequence of joint angles and control actions. 
Due to unmodelled effects such as hysteresis or unknown contacts, the state transitions are non-Markovian even if joint angles and velocities are known.
We can assume that the observations are almost noise-free since the measurement errors for our observations (joint angles) are minimal.
Nevertheless, as the observations do not contain the full state information of the system, we still have to model uncertainty in our latent state using the RKN. 
\begin{wrapfigure}[21]{l}{0.45\linewidth}
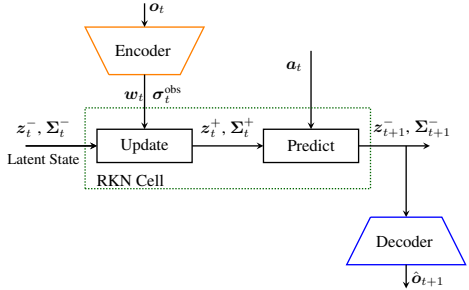


%      \includegraphics[scale=0.5]{Figures/RKN-SchematicArrow.pdf}
 %     \caption{Schematic Diagram Of Action Conditional Recurrent Kalman Network. Action conditioning takes place via additive interactions between the latent state $\vec{z}_{t}$ and action $a_t$ at the predict stage in the RKN Cell.}

%     \hspace*{0.2cm} 
   %   \includegraphics[scale=0.4]{Figures/ForwardArrow.pdf}
   \begin{subfigure}[b]{0.45\textwidth}
     \centering
     \resizebox{\textwidth}{!}{\tikzACRKN}
      %\caption{}

     \end{subfigure}

\caption{ Schematic diagram of ac-RKN for forward dynamics learning. Action conditioning is implemented by adding a latent control vector $\vec b(\vec a_t)$ to the dynamics model of the RKN. The output of the predict stage, which forms the prior for the next time step $\left(\vec{z}_{t+1}^-, \vec{\Sigma}_{t+1}^- \right)$ is decoded to get the prediction of the next observation. \label{fig:forward_model}}

\end{wrapfigure}
    
The function $f$ can be challenging to learn when the observations, i.e., joint positions and velocities, of two subsequent time-steps are too similar.
In this case, the action has seemingly little effect on the output and the learned strategy is often to copy the previous state for the next state prediction.
This difficulty becomes more pronounced as the time step between states becomes smaller ,e.g., $1$ms, as minor errors in absolute states estimates can already create unrealistic dynamics. 
Hence, a standard method for model learning is to predict the normalized differences between subsequent observations or states instead of the absolute values~\cite{deisenroth2011pilco}, i.e., during training, the predicted next observation is
%\hat{\vec o}_{t+1} = \vec o_t + f( \vec o_{1:t}, \vec a_{1:t} )$, where $\hat{\vec o}_{t+1}$ is the predicted next observation. 
$\hat{\vec o}_{t+1} = \vec o_t + \textrm{dec}\left(\vec{z}_{t+1}^-\right)$, where $\vec{o}_t$ is the true observation at $t$ and  $\textrm{dec}\left(\vec{z}_{t+1}^-\right)$ is the  output of the actual decoder network. 
%We extend this to non-Markovian models by making the decoder predict the differences to next observation.
%In case of missing observations, the probabilistic formulation of the recurrent uses the prior for the prediction.
During inference, we introduce an additional memory $\vec{m}_t$ which stores the last prediction and uses it as observation in case of a missing observation, i.e.,
$\vec{\hat{o}}_{t+1} = \vec{m}_t + \textrm{dec}\left(\vec{z}_{t+1}^-\right)$ where we use the true observation, $\vec{m}_t= \vec{o}_t$ if available and the predicted observation, $\vec{m}_t= \hat{\vec{o}}_t$, otherwise. 
The architecture of the ac-RKN is summarized in Figure \ref{fig:forward_model}.

The RKN cell is well equipped to handle missing observations, as the update step can just be omitted in this case, i.e., the posterior is equal to the prior if no observation is available.
%due to its probabilistic nature. 
Thus, we can also use the presented architecture to predict several time steps into the future.
In this case, we repeatedly apply the RKN prediction step and update the memory $\vec m_t$ with the corresponding prediction while omitting the Kalman update.

\paragraph{Loss And Training.}

%If uncertainty estimates of the predictions are required we use the negative Gaussian log-likelihood(NLL),  $\mathcal{L}_\textrm{nll}$, otherwise the root mean squared error (RMSE) $\mathcal{L}_\textrm{rmse}$ is sufficient. Let $\vec{o}_{1:T}$ be the ground truth observation sequence with dimension $D_s$. Then the negative Gaussian log-likelihood is computed by 
%\begin{align}
%\mathcal{L}_\textrm{nll} =  -
%\dfrac{1}{T} \sum_{t=1}^T  \log \mathcal{N}\left(\vec{o}_{t+1} \bigg| \vec m_t + \textrm{dec}_{\mu}(\vec{z}^-_{t+1}), \textrm{dec}_{\Sigma} (\vec{\sigma}^-_{t+1})\right),
%\end{align}
%where $\textrm{dec}_{\mu}$ and $\textrm{dec}_{\Sigma}$ denote the parts of the decoder that are responsible for decoding the latent mean and latent covariance respectively.%, and $\vec m_t$ is the memory of the difference predictor, i.e., the old observation in most cases.
%where $\vec m_t$ is the memory of the difference predictor, i.e., the old observation in most cases, and  $\textrm{dec}_{\mu}$ and $\textrm{dec}_{\Sigma}$ denote the parts of the decoder that are responsible for decoding the latent mean and latent variance respectively.
For training forward dynamic models, we optimize the root mean square error (RMSE) between the decoded output and normalized differences to the the next true state $\left(\vec{o}_{t+1} -  \vec{o}_{t}\right)$ as in \cite{nagabandi2018neural},    
\begin{align*}
\mathcal{L}_{\textrm{fwd}} = \sqrt{
\dfrac{1}{T}\sum_{t=1}^T  \parallel \left(\vec{o}_{t+1} -  \vec{o}_{t}\right) - \textrm{dec}\left(\vec{z}^-_{t+1}\right) \parallel^2}.
\end{align*}
Note that we decode the prior mean $\vec{z}^-_{t+1}$ for predicting $\vec o_{t+1}$. 
The prior mean $\vec{z}^-_{t+1}$ integrates all information up to time step $t$, including $\vec a_t$, but already denotes the belief for the next time step. 
Hence, as we are interested in prediction, we work with this prior.
In contrast, \cite{becker2019recurrent} used the posterior belief since their goal was filtering rather than prediction. 

Further, in \cite{becker2019recurrent}, the Gaussian log-likelihood is used instead of the RMSE. 
While this is in principle also possible here, it is only necessary if uncertainty estimates in the outputs are required. 
If this is not the case, training on RMSE yields slightly better predictions and allows for a fair comparison with deterministic baselines like feed-forward neural networks, LSTMs and analytical models. 
Thus, in this work, we chose to work with the RMSE loss.

%If we do not require uncertainty estimates, training on the RMSE yields slightly better predictions. % in terms of the squarred error. 
%Therefore, we evaluate on the root mean squared error in order to have a fair comparison with the deterministic models like feed forward neural networks and LSTMs. 
%The architecture of the ac-RKN is summarized in Figure \ref{fig:forward_model}.

\subsection{Inverse Dynamics Learning}
\label{sec:inverseDyn}
For the inverse dynamics case  we want to learn a model \mbox{$f^{-1}: \vec o_{1:t}, \vec a_{1:{t-1}}, \vec o_{t+1} \mapsto \hat{\vec{a}}_{t}$} where $\vec o_{t+1}$ is the desired next observation for time step $t + 1$ and $\hat{\vec{a}}_{t}$ is the predicted action, i.e., the one to be applied. We introduce an action decoder which decodes the latent posterior $\left(\vec{z}_{t}^+, \vec{\Sigma}_{t}^+ \right)$ and estimates the action required to move to the desired next observation. The action decoder also gets information regarding the next observation as an input.
During training this corresponds to the next observation in the data. 
For control a desired next observation is used to obtain the action required to reach that observation. The joint angles and velocities are treated as observations in our inverse dynamics experiments.
\begin{wrapfigure}[22]{r}{0.45\linewidth}
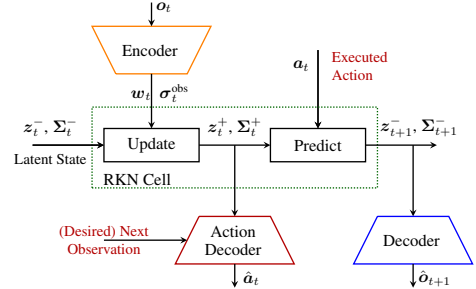

     \begin{subfigure}[b]{0.45\textwidth}
     \centering
     \resizebox{\textwidth}{!}{\tikzACRKNINVDYN}
      \end{subfigure}
\caption{Schematic diagram of the inverse dynamics learning architecture. Here the posterior $\left(\vec{z}_{t}^+, \vec{\Sigma}_{t}^+ \right)$, at the current time step is fed to an action decoder along with the desired target. In the next time-step, the executed action $\vec a_t$ is fed to the predict stage of RKN cell. 
Further, the next predicted prior $\left(\vec{z}_{t+1}^-, \vec{\Sigma}_{t+1}^- \right)$ is decoded to get the next state $\hat{\vec{o}}_{t + 1}$. \label{fig:inverse}}
\end{wrapfigure}

As seen in Figure \ref{fig:inverse}, instead of merely learning the inverse dynamics, our approach learns inverse and forward dynamics simultaneously by feeding back the executed action to the action-conditional predict stage, which predicts forward in latent space. 
%Note that the executed action is not necessarily the same as the one predicted by the model.
%Yet, the predicted action $\vec{\hat{a}}_t$ is not necessarily the same as the one executed. We therefore encounter two different situations where the prediction step gets: (i) the predicted action $\hat{\vec a}_t$ as input or (ii) the real executed action $\vec{a_t}$ as input. 
%Clearly, (ii) is the more reasonable model as observing the real executed actions contains more information about the dynamics of the robot than the predicted actions which were never executed and hence this is feed to the ac-RKN. 
Note that the action decoder also gets the same action as the target during training.
However, the model sees the true action only after the prediction since the action decoder works with the posterior estimate $\left(\vec{z}_t^+, \vec{\Sigma}_t^+ \right)$.  
%Note that unlike the decoder in the forward dynamics model, we decode the posterior $\left(\vec{z}_t^+, \vec{\Sigma}_t^+ \right)$ to predict the action $\hat{\vec{a}}_t$. 
%Thus, as the action decoder has to predict based on the posterior estimate, the model sees the action only after it has made its prediction.
Hence, the prediction of the inverse dynamics model for the current time step is made independent of this feedback.
We found that enforcing this causal feedback, a necessary structural component of the Bayesian network of the underlying latent dynamical system, improves the performance of the inverse model, as seen in Figure \ref{fig:franka_inter}.
% One might argue that a comparison to standard methods is unfair as our approach gets the real control actions also as input, so it might exploit the temporal correlations in this signal. For this reason, we also evaluated (a), where the network does not have access to the ground-truth actions.  Note that in (a) the output of the action decoder is connected to action conditional input at the predict stage during both training as well as inference. This connection allows the flow of additional analytic gradients through the learned dynamics which can have interesting implications in several applications. 

%The final torque ($A_t$) applied to the robot can be the predicted actions as it is or in conjunction with a feedback controller. $A_t$ is also used to make action conditional predict step(transition model) of the RKN. As in Figure (1) the latent state is also decoded after the predict step of the RKN to obtain the next state $s_{t+1}$ to learn an implicit forward model which helps in regularzing the inverse model.

\paragraph{Loss And Training.}
The dual output architecture for our inverse model leads to two different loss functions for the action and observation decoders which are optimized jointly.
Our experiments showed that the loss of the forward model is an excellent auxiliary loss function for the inverse model and learning this implicit forward model jointly with the inverse model results in much better performance for the inverse model. 
We assume the reasons for this effect is that the forward model loss is providing crucial gradient information to form an informative latent state representation that is also useful for the inverse model.
%We can again choose between the negative Gaussian log-likelihood loss and the RMSE, depending on whether we require uncertainty estimates or not. 
In order to deal with high frequency data we again chose to predict the normalized differences to the previously executed action, i.e. $\hat{\vec{a}}_t = \vec{a}_{t-1} + \textrm{dec}_\textrm{action}(\vec{z}_t^+)$. 
Here $\vec{a}_{t-1}$ is the executed action at $t-1$ and $\textrm{dec}_\textrm{action}(\vec{z}_t^+)$ the output of the actual action decoder network. 
The combined loss function for the inverse dynamic case is given by 

%\begin{align}
%\begin{split}
%\mathcal{L}_{\textrm{inv}} = 
%\sqrt{\dfrac{1}{T}  {\sum_{t=1}^T  %\big\lVert{\bar{\vec{a}}_{t+1}- %\textrm{dec}_{\mu_a}(\vec{z}^+_{t})}\big\rVert^2}}  \quad  + 
% \quad \large{\lambda} \sqrt{
%\dfrac{1}{T} {\sum_{t=1}^T  \big\lVert{\bar{\vec{o}}_{t+1} - \textrm{dec}_{\mu}(\vec{z}^-_{t+1})}\big\rVert^2}}, 
%\end{split}
%\label{eq:invLoss}
%\end{align}
 \begin{align*}
     \mathcal{L}_\textrm{inv} = \sqrt{\dfrac{1}{T} \sum_{i=1}^T \parallel \left(\vec{a}_{t+1} - \vec{a}_{t}\right) - \textrm{dec}_\textrm{action}\left(\vec{z}_t^+\right) \parallel^2}    + 
     \lambda \mathcal{L}_\textrm{fwd} %\sqrt{ \dfrac{1}{T} \sum_{t=1}^T  \parallel \vec{o}_{t+1} - \left( \vec{o}_{t} + \textrm{dec}\left(\vec{z}_t^- \right)  \right)
     %\parallel^2 
 \end{align*}
where 
%$\bar{\vec{a}}_{t+1} = \vec{a}_{t+1} - \vec{a}_{t}$, $\bar{\vec{o}}_{t+1} = \vec{o}_{t+1} - \vec{o}_{t}$ and
$\lambda$ chooses the trade-off between the inverse model loss and the forward model loss. The value of $\lambda$ is chosen via hyperparameter optimization using GPyOpt~\cite{gpyopt2016}.
Note that for the inverse model,  the actions are decoded based on the posterior mean of the current time step while for the forward model the observations are decoded from the prior mean of the next time step. 

%where $\vec{\hat{o}}_{t}$ and $\vec{\hat{\sigma}}_{t}$ denote the parts of the decoder that are responsible for decoding the latent mean and latent variance respectively.\\

%As in forward dynamics learning  gradients are computed using (truncated) backpropagation through time (BPTT) \cite{werbos1990bptt} and clipped. The Adam \cite{kingma2014adam} stochastic gradient descent optimizer with default parameters is used for optimization.
\section{Experimental Results}
\label{sec:result}

In this section, we discuss the experimental evaluation of ac-RKN on learning the forward and inverse dynamics models on robots with different actuator dynamics. A full listing of hyperparameters can be found in the supplementary material. We compare ac-RKN to both standard deep recurrent neural network baselines (LSTMs, GRUs and standard RKN), analytical baselines based on classical rigid body dynamics and non-recurrent baselines like FFNN. For RKN and LSTM baselines, we replaced the ac-RKN transition layer with generic LSTM and RKN layers. For recurrent models, the recurrent cell parameters are tuned via hyperparameter optimization using  GPyOpt~\cite{gpyopt2016}, but the encoder and decoder sizes are similar. The observations and actions are concatenated as in \cite{nagabandi2018neural} and \cite{deisenroth2011pilco} for all baseline experiments during the forward dynamics learning, i.e., we treat actions as extended observations. The data, i.e., the states or observations, actions and targets are normalized (zero mean and unit variance) before training. We denormalize the predicted values during inference and evaluate their performance on the test set. We evaluate the model using RMSE to ensure a fair comparison with the deterministic models such as FFNNs and LSTMs. 

\subsection{Forward Dynamics Learning}
\label{sec:fwd_dyn_res}
 \begin{figure*}[t]
\hspace*{0.9cm}

\begin{minipage}{\textwidth}
\begin{subfigure}{.33\textwidth}
\vspace{0.5cm}
\begin{tabular}{|l|c|c|} \hline
         & RMSE  & NLL \\ \hline
ac-RKN &\textbf{0.0144} & \textbf{6.247}    \\ \hline
RKN    & 0.0282 & 4.930     \\ \hline
LSTM &0.2067  & 0.952\\ \hline
GRU &0.2015  & 1.186\\ \hline
\end{tabular}
\vspace{0.5cm}
\caption{Hydraulic Brokk 40}

  \end{subfigure} 
  \begin{subfigure}{.32\textwidth}
    \includegraphics[width=\linewidth]{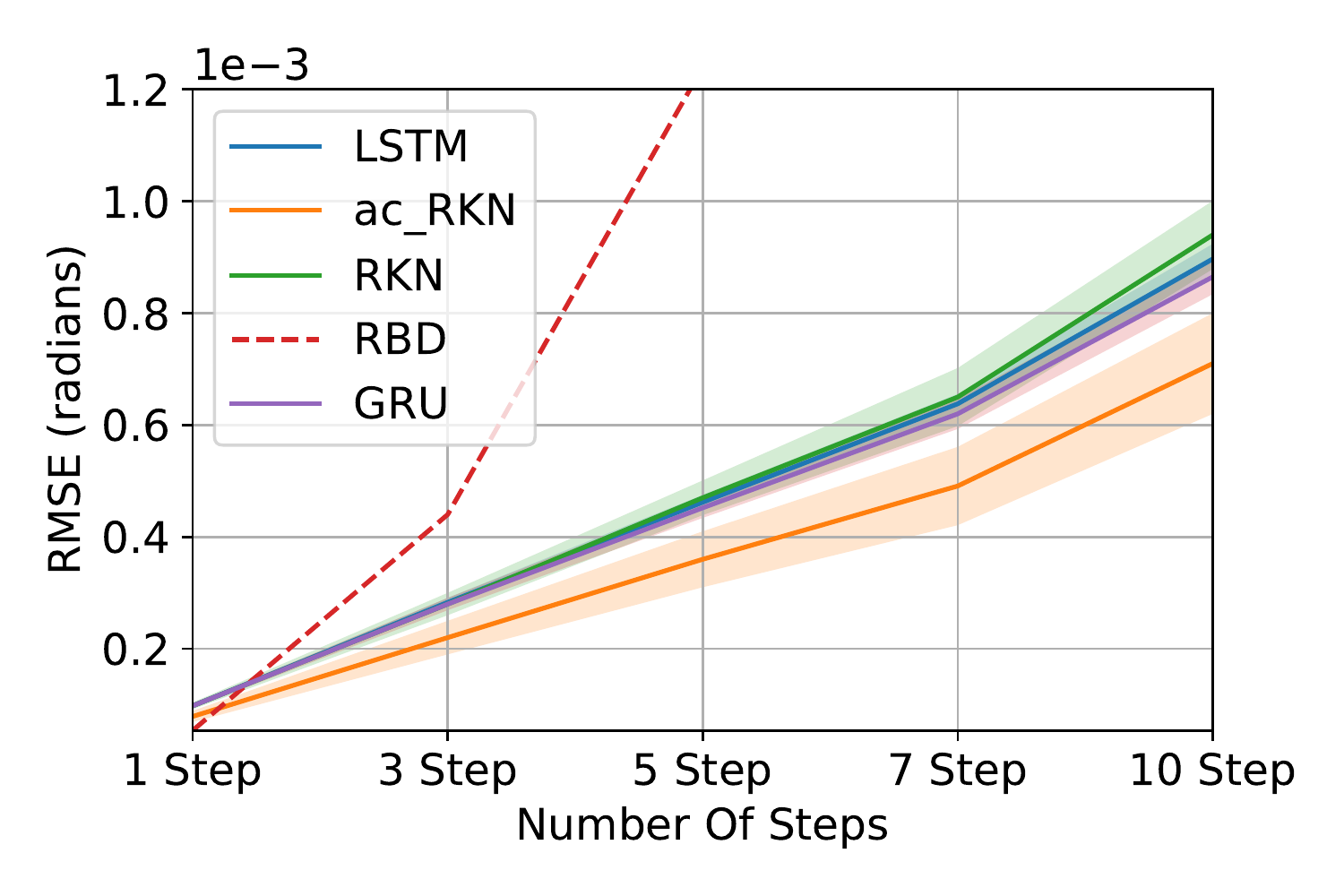}
    \caption{Franka Emika Panda}
  \end{subfigure} 
  \begin{subfigure}{.32\textwidth}
    \includegraphics[width=\linewidth]{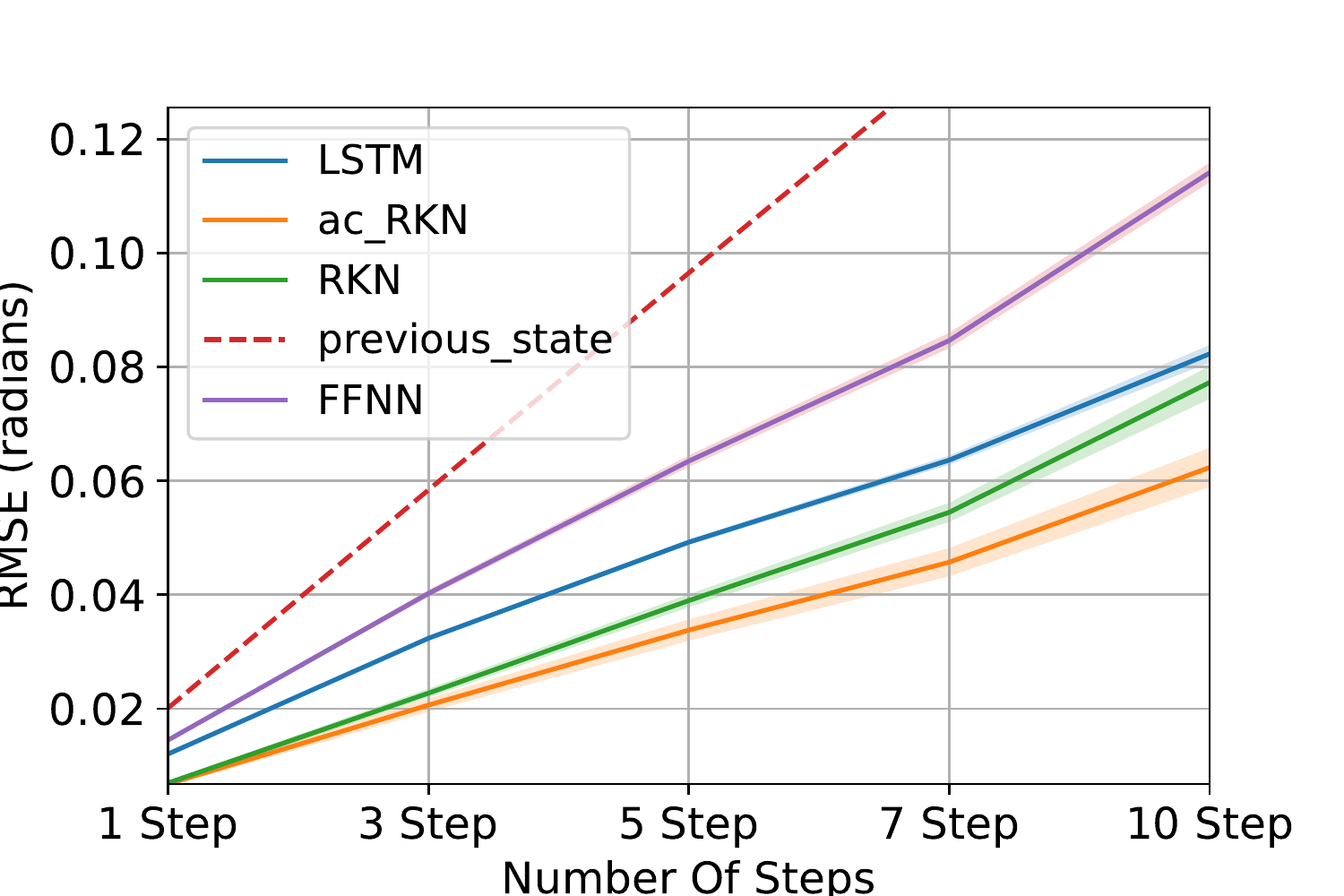}%\\{\color{}}
    \caption{Pneumatic Arm}
  \end{subfigure}

\vspace{0.002\textwidth}
\end{minipage}

\vspace{0.1cm}  
\caption{\textbf{(a)} Performance comparison of various recurrent models for the BROKK-40 hydraulically actuated robot arm. \textbf{(b)} and \textbf{(c)} Comparison of multi-step ahead prediction for different algorithms for Franka Emika and the pneumatically actuated muscular robot arm. The shaded region shows standard deviations. The dashed line in (b) denotes the prediction performance of the rigid body dynamics model and in (c) it denotes the prediction performance if we just predict the previous state. The results show that all learning approaches outperform the rigid body dynamics model or the previous state prediction, while our ac-RKN architecture outperformed other learning methods.   \label{fig:forwardModelPerformance}
}
%\vspace{-0.5cm}
\end{figure*}  
\begin{figure*}[t]
\hspace*{-0.1cm}

%\begin{minipage}{\textwidth}
\centering
\begin{subfigure}{.32\textwidth}
\vspace{0.3cm}
    \includegraphics[width=\linewidth]{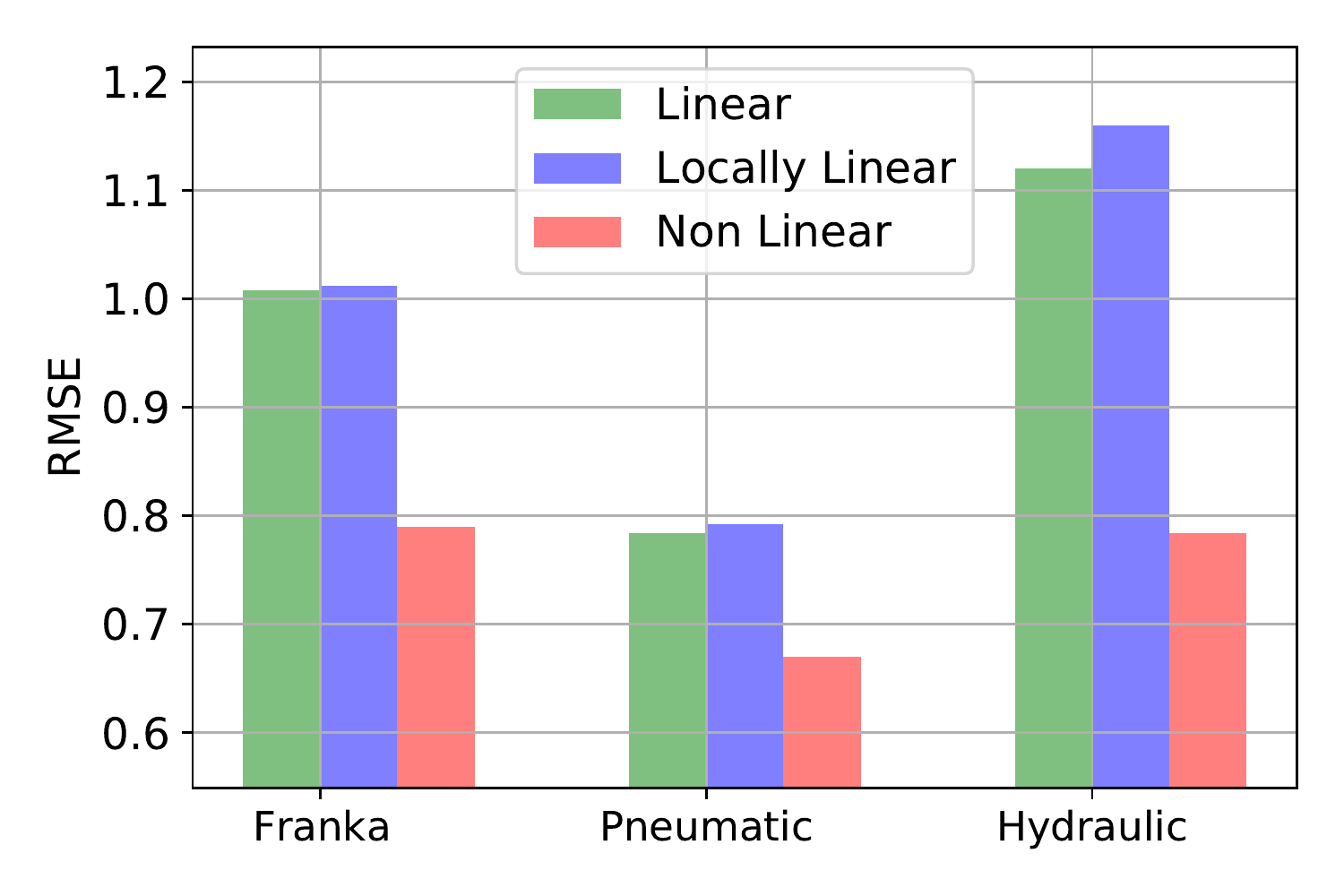} 
    \vspace{-0.3cm}
    \caption{\small Impact of control input models}
    \label{fig:control}
  \end{subfigure}
   \begin{subfigure}{.29\textwidth}
    \includegraphics[width=\linewidth]{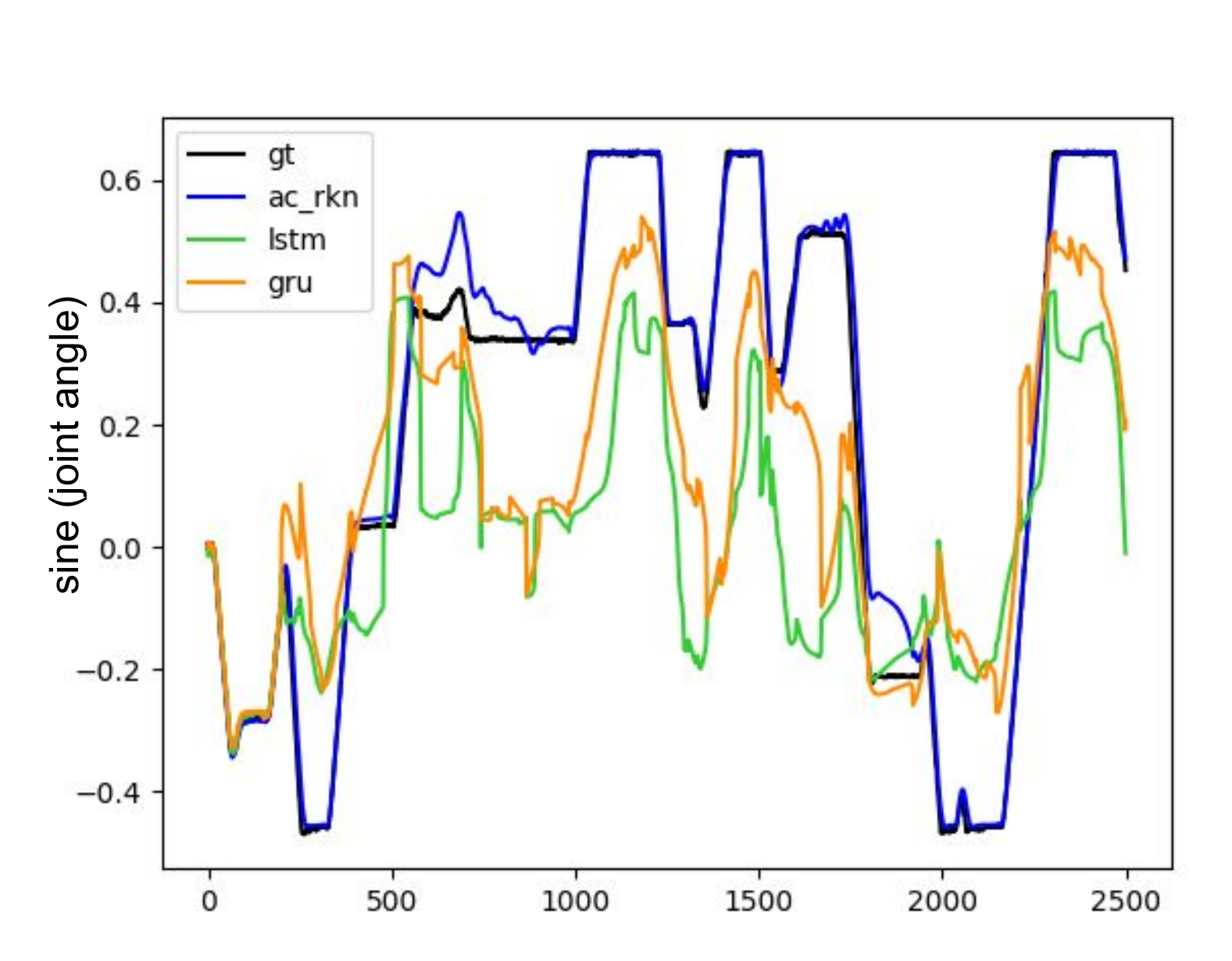}%\\{\color{}}
    \vspace{0.2cm}
    \caption{Hydraulic Arm Predictions}
    \label{fig:hyd_traj}
  \end{subfigure}
\begin{subfigure}{.32\textwidth}
    \includegraphics[width=\linewidth]{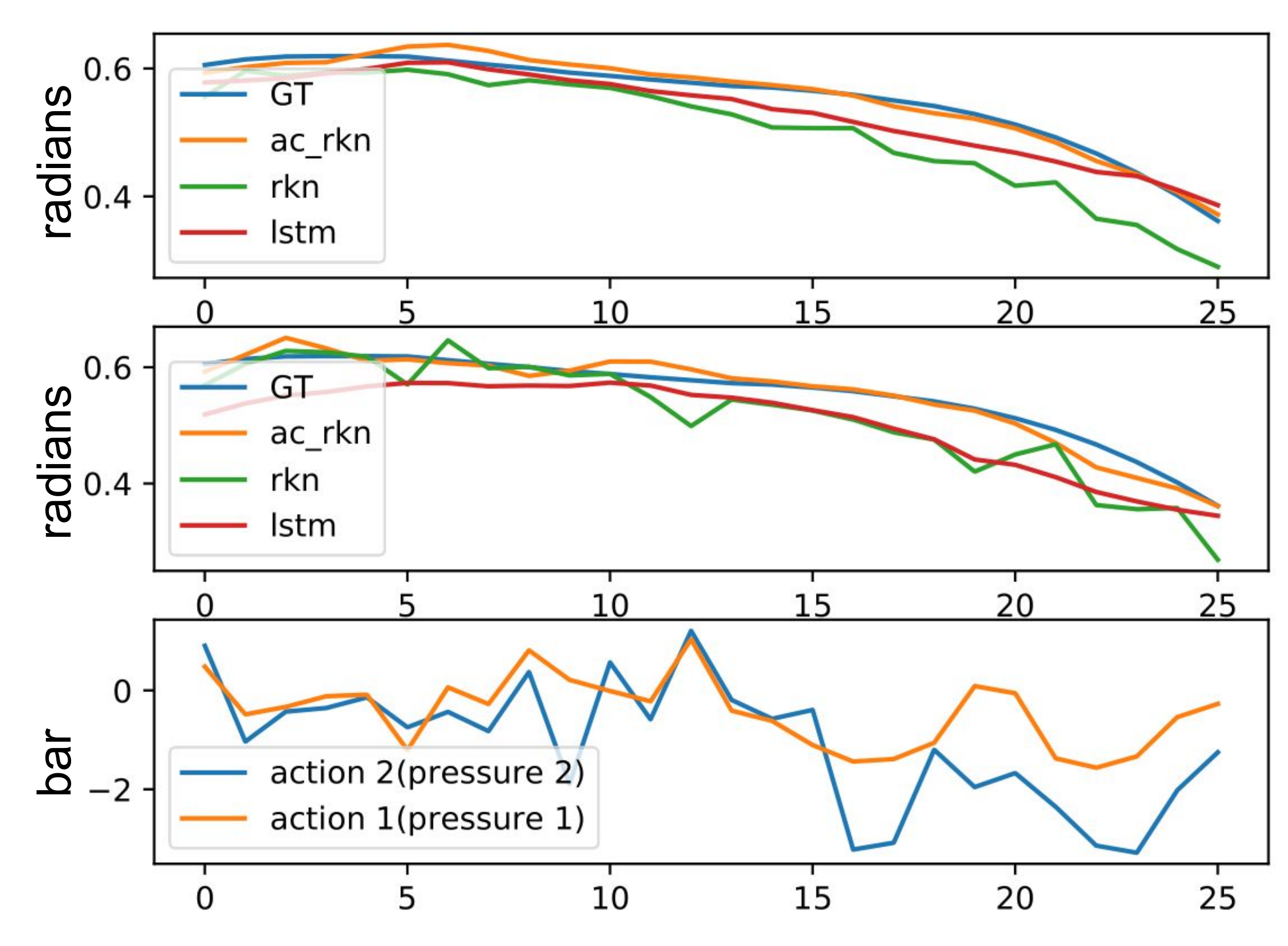}
    \caption{Pneumatic Arm Predictions}
    \label{fig:pam_traj}
  \end{subfigure}
%\vspace{0.002\textwidth}
%\end{minipage}

%\vspace{0.1cm}  
\caption{\textbf{(a)} Comparison of different action-conditioning models discussed in \ref{sec:action-con}. \textbf{(b)} Visualization of the predicted trajectories of the hydraulic arm for 200 step ahead predictions. \textbf{(c)} Visualization of the predicted trajectories of the pneumatic arm for 5 (top) and 10 (bottom) step ahead predictions for one of the joints. (b) and (c) clearly show that the predictions given by our approach are by far the most accurate. \label{fig:forwardModelTrajectories}
}
\vspace{-0.5cm}
\end{figure*}  

We evaluate the ac-RKN on three robotic systems with different actuator dynamics each of which poses unique learning challenges:  \textbf{(i) Hydraulically Actuated BROKK-40 Robot Arm.} The data consists of measured joint positions and the input current to the controller sampled at 100Hz from a hydraulic BROKK-40 demolition robot~\cite{taylor2013state}. The position angle sensors are rotary linear potentiometers. We chose a similar experimental setup and metrics as in \cite{becker2019recurrent} to ensure a fair comparison. We trained the model to predict the joint position 2 seconds, i.e.  200 time-steps, into the future, given only control inputs. Afterwards, the model receives the next observation and the prediction process repeated. Learning the forward model here is difficult due to inherent hysteresis associated with hydraulic control. 
\textbf{(ii) Pneumatically Actuated Musculoskeletal Robot Arm.} This four DoF robotic arm is actuated by Pneumatic Artificial Muscles (PAMs)~\cite{buchler2016lightweight}. 
Each DoF is actuated by an antagonistic pair of PAMs, yielding a total of eight actuators.
The robot arm reaches high joint angle accelerations of up to 28,000 $\textrm{deg}/s^2$ while avoiding dangerous joint limits thanks to the antagonistic actuation and limits on the air pressure ranges. 
The data consists of trajectories of hitting movements with varying speeds while playing table tennis~\cite{buchler2020learning} and is recorded at 100Hz. The fast motions with high accelerations of this robot are complicated to model due to hysteresis. 
\textbf{(iii) Franka Emika Panda Arm.}  We collected the data from a 7 DoF Franka Emika Panda manipulator during free motion at a sampling frequency of 1kHz.
It involved a mix of movements of different velocities from slow to swift motions. 
The high frequency, together with the abruptly changing movements results in complex dynamics which are interesting to analyze.

\textbf{Multi Step Ahead Prediction}. Figure \ref{fig:forwardModelPerformance} summarizes the test set performance for each of these robots.  We benchmarked the performance of ac-RKN with state of the art deterministic deep recurrent models (LSTM and GRU) and non-recurrent models like feed-forward neural network (FFNN). In all three robots, the ac-RKN gave much better multi-step ahead prediction performance than these deterministic deep models. The performance improvement was more significant for robots with hydraulic and pneumatic actuator dynamics due to explicit non-markovian dynamics and hysteresis, which is often difficult to model via analytical models. We also validated the performance improvement due to our principled action-conditioning in the latent transition dynamics by comparing it with RKN~\cite{becker2019recurrent}, which treats actions as part of the observations by `concatenation'. In all three robots, our principled treatment brought significant improvement in performance. Figure \ref{fig:forwardModelTrajectories} (b,c) shows the predicted trajectories for the BROKK and the pneumatically actuated robot arm.

\textbf{Comparison with Analytical Model}. We were also interested in making a comparison with analytical models of Franka. For the analytical model, aside from the inertia properties of the links, the Coulomb friction was also identified for every joint. A detailed description for the same can be found in Appendix C. As seen in Figure \ref{fig:forwardModelPerformance} (b), the performance of the analytical model outperformed ac-RKN for one step ahead predictions, but for multi-step forward predictions, the data-driven models had a clear advantage with ac-RKN giving the most accurate results.

\textbf{Ablation Study for Action-Conditioning.}
We evaluated different models for action conditioning as discussed in Section \ref{sec:forwardDyn}. The resulting evaluation can be seen in Figure \ref{fig:forwardModelTrajectories}(a) and shows the advantage of using non-linear models for the additive action-conditioning of the latent dynamics.
%Thus we believe ac-RKN can be a viable data-driven model for model-based control of large hydraulic robots and flexible pneumatic soft robots. 

\subsection{Inverse Dynamics Learning}

%\textbf{Robotic Platforms}.
We evaluated the performance of the proposed method for inverse dynamics learning on two real robots, Franka Emika Panda and Barrett WAM. Barret WAM is a robot with direct cable drives. The direct cable drives produce high torques, generating fast and dexterous movements but yield complex dynamics. Rigid-body dynamics cannot model this complex dynamics accurately due to the variable stiffness and lengths of the cables~\cite{nguyen2011incremental}. 
 \begin{figure*}[t]
\hspace*{-1cm}
%\noindent\begin{minipage}{\textwidth}
%\begin{minipage}{\textwidth}
\centering
  \begin{subfigure}{.34\textwidth}
    \includegraphics[width=\linewidth]{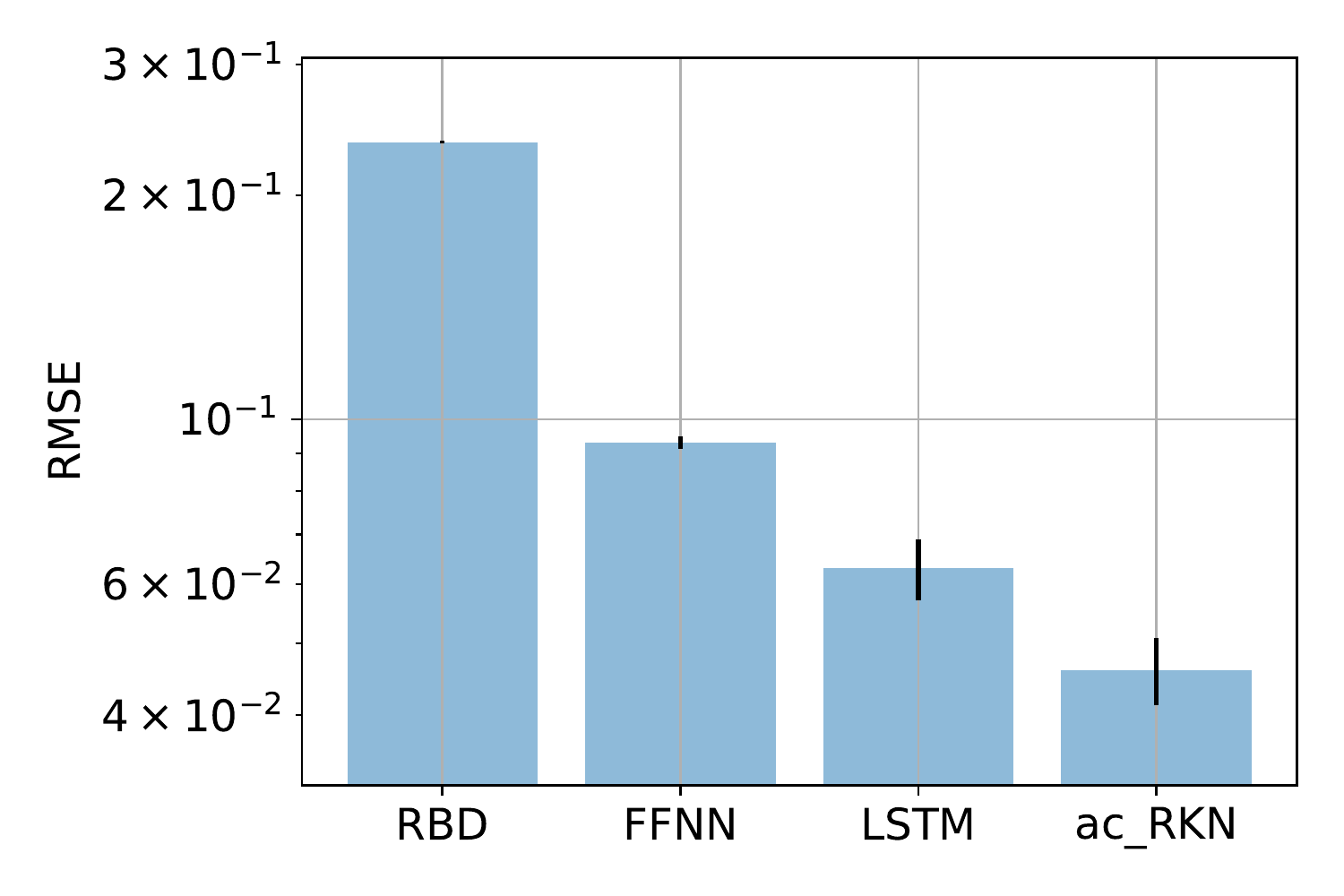}
    \caption{Franka Panda}
    \label{fig:frnaka_inv_torque}
  \end{subfigure} 
%   \qquad
%   \begin{minipage}{.32\textwidth}
%     \includegraphics[width=\linewidth]{Figures/pamForward.pdf}%\\{\color{}}
%   \end{minipage}
\begin{subfigure}{.34\textwidth}
    \includegraphics[width=\linewidth]{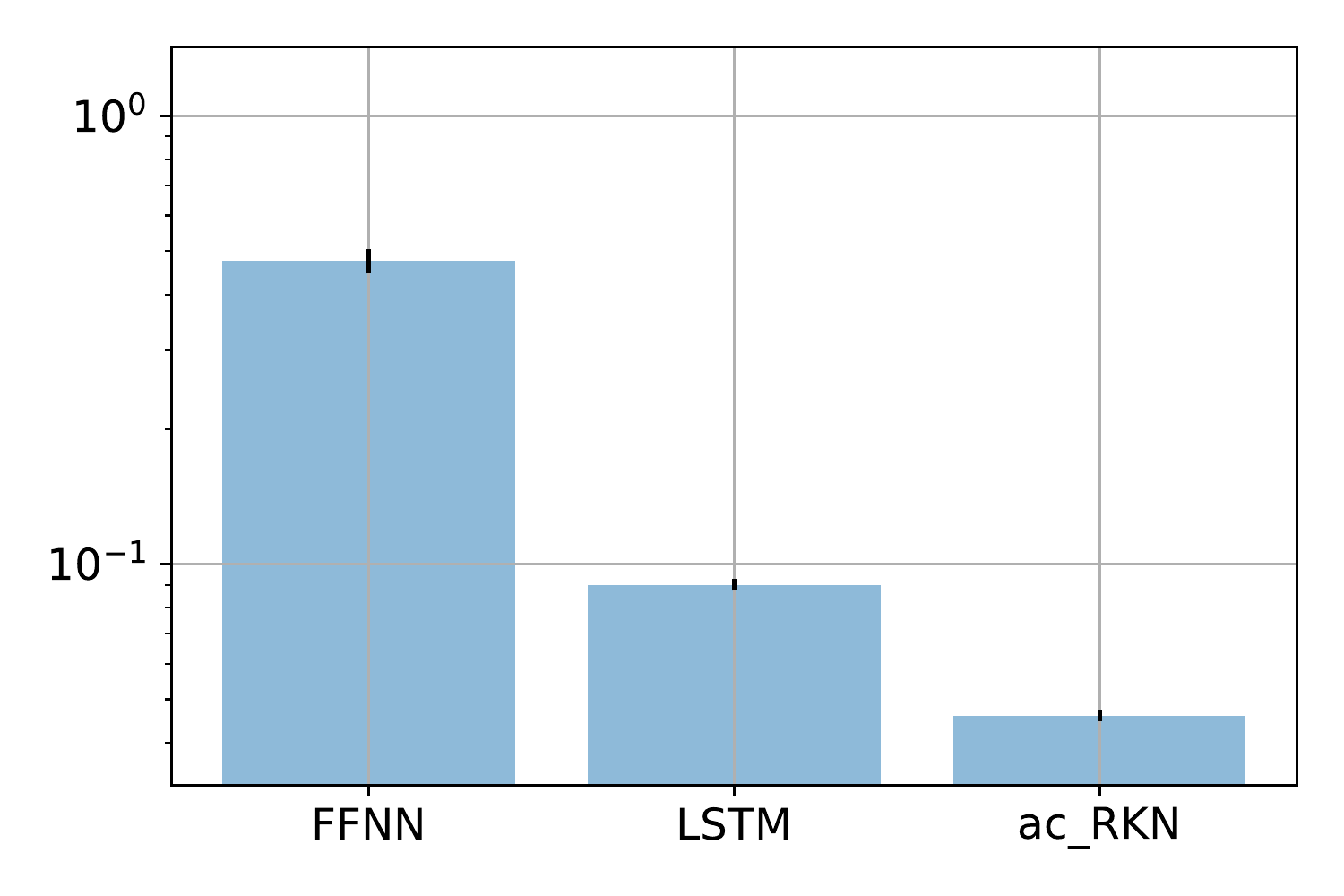}
    \caption{Barret WAM}
    \label{fig:barett_inv}
  \end{subfigure}
    \begin{subfigure}{.23\textwidth}
    \includegraphics[width=\linewidth]{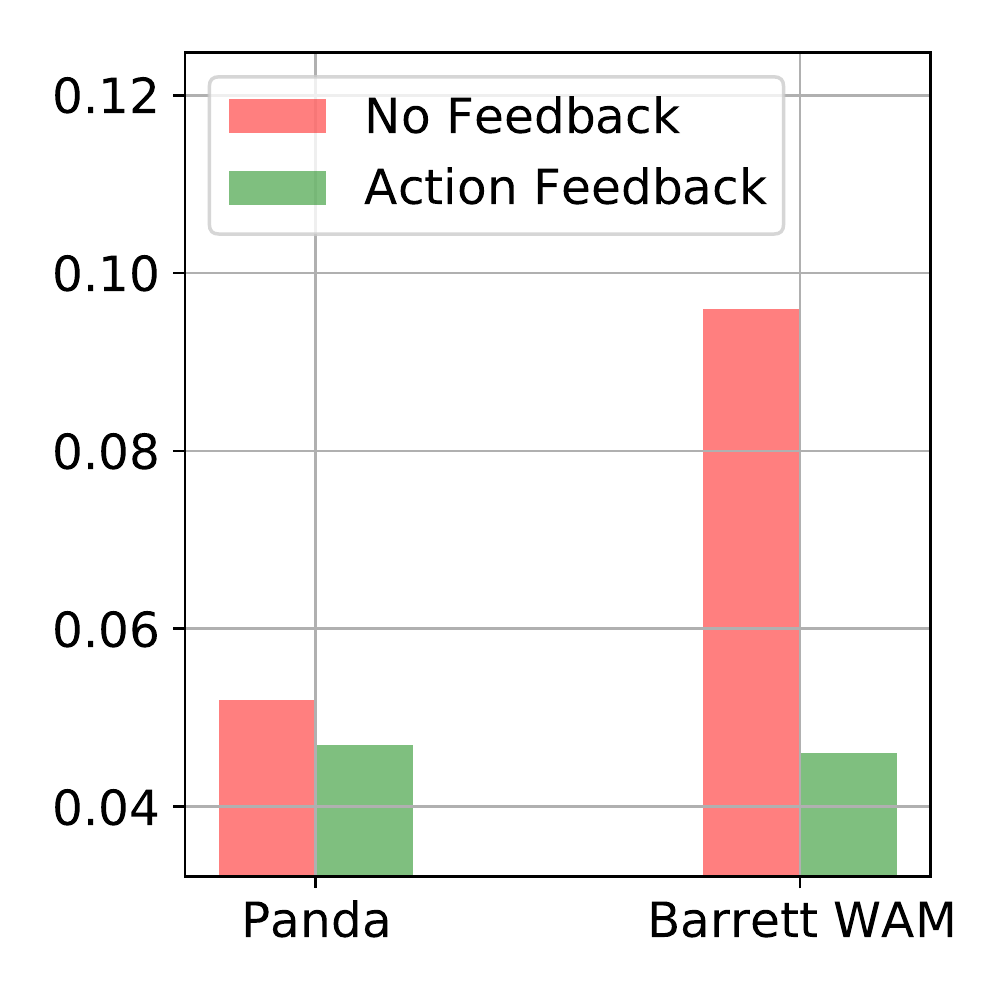}
    \caption{Action Feedback}
    \label{fig:franka_inter}
  \end{subfigure}
%   \begin{subfigure}{.24\textwidth}
%     \includegraphics[width=\linewidth]{Figures/barret_causal_inv.pdf}
%     \caption{}
%     \label{fig:barett_inter}
%   \end{subfigure}
%\vspace{0.002\textwidth}
%\end{minipage}
%\vspace{0.002\textwidth}
%\end{minipage}
%\vspace{-0.2cm}  
\caption{\textbf{(a)} and \textbf{(b)} Joint torque prediction RMSE values in NM of Action-Conditional RKN, LSTM and FFNN for Panda and Barret WAM. A comparison is also provided with the analytical (RBD) model of Panda. \textbf{(c)} Comparison of ac-RKN for inverse dynamics learning with and without the action feedback as discussed in Section \ref{sec:inverseDyn}.
}
\vspace{-0.5cm}
\end{figure*}  

\begin{figure*}[t]
\centering
\hspace*{0.2cm}
%\noindent\begin{minipage}{\textwidth}
%\begin{minipage}{\textwidth}
\begin{minipage}{.33\textwidth}
\includegraphics[width=\textwidth]{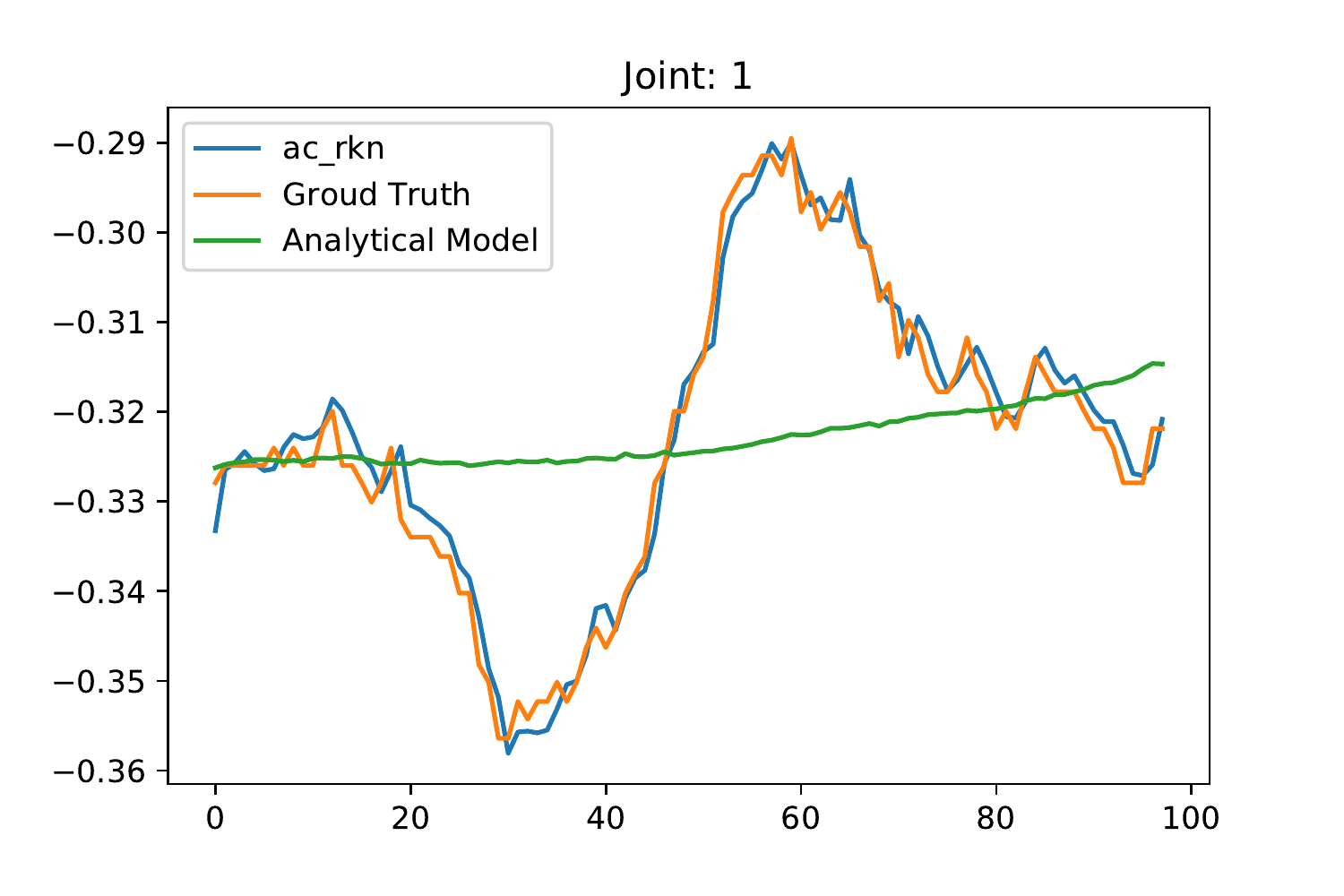}
\end{minipage}% 
\begin{minipage}{.33\textwidth}
\includegraphics[width=\textwidth]{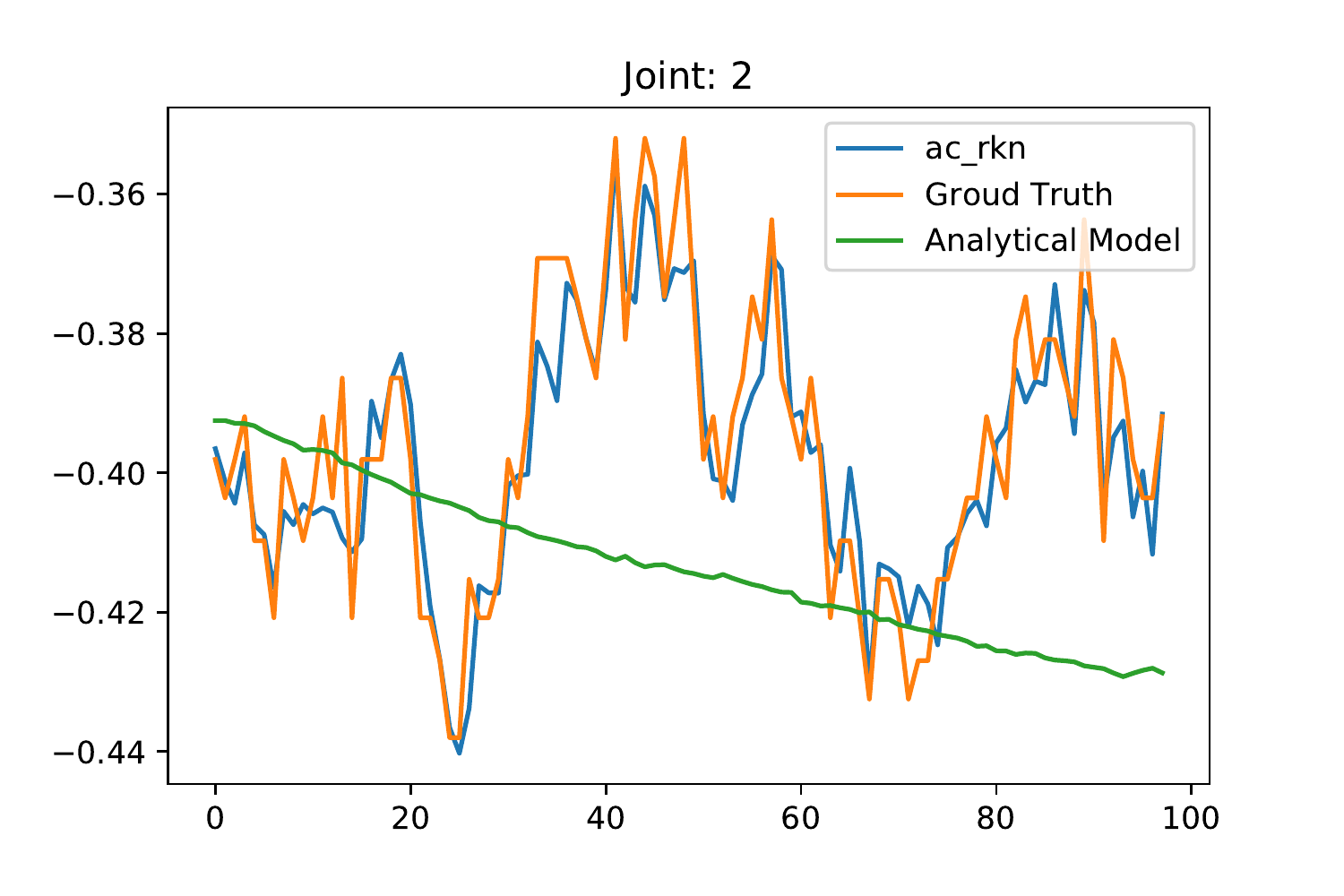}
\end{minipage}%
\begin{minipage}{.33\textwidth}
\includegraphics[width=\textwidth]{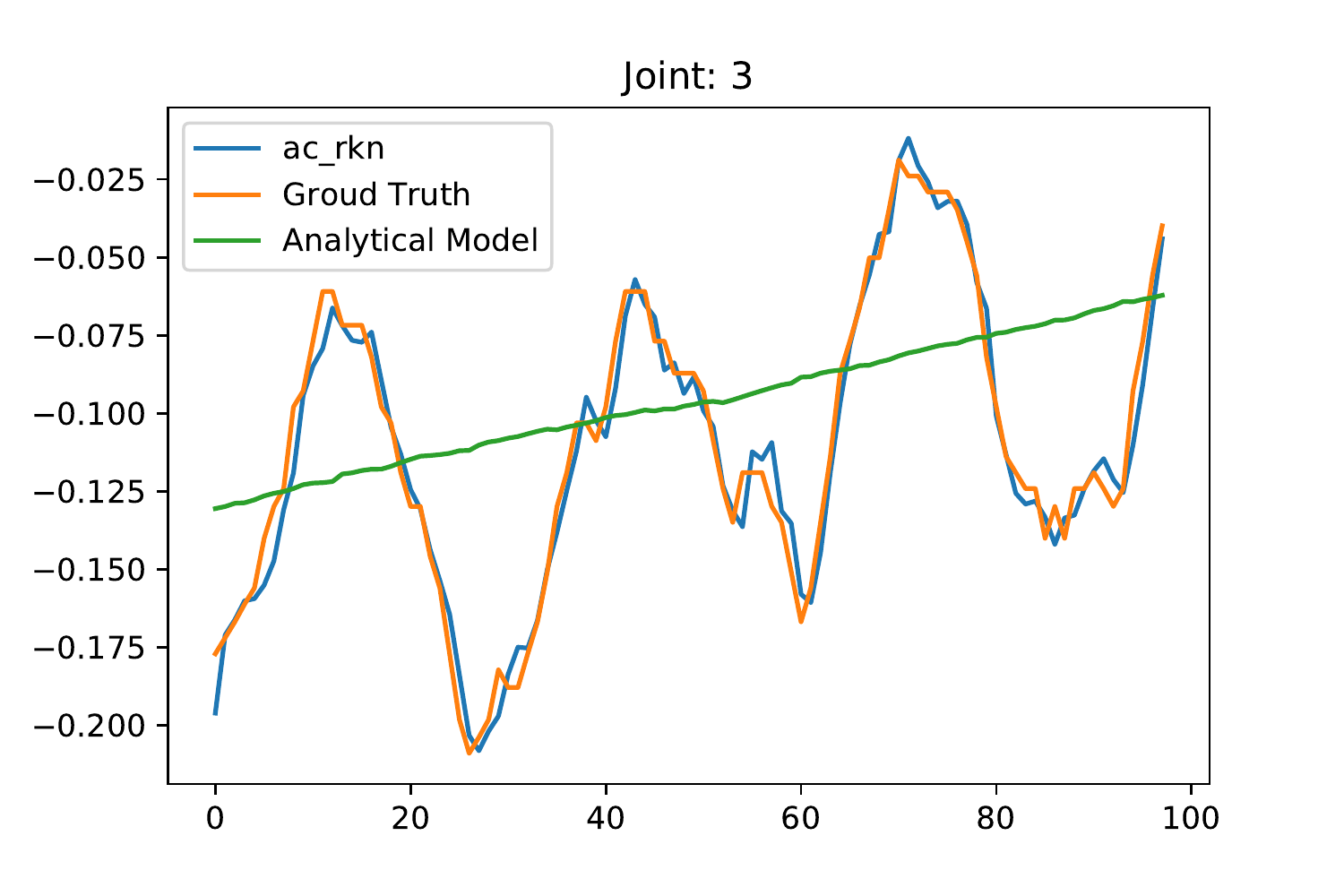}
\end{minipage}
  % \begin{minipage}{.24\textwidth}
  %  \includegraphics[width=\textwidth%]{Figures/joint_4.pdf}
  %\end{minipage}
%   \begin{minipage}{.19\textwidth}
%     \includegraphics[width=\linewidth]{Figures/joint_5.pdf}
%   \end{minipage}
%\vspace{0.002\textwidth}
%\end{minipage}

%\vspace{0.002\textwidth}
%\end{minipage}
%\vspace{0.1cm}  
\caption{ Predicted joint torques(normalized) for first 3 joints of the Panda robot arm. The learned inverse dynamics model by ac-RKN match closely with the ground truth data while the rigid body dynamics model can not capture the high-frequency variations in the data. 
}
\label{fig:inv_analytical}
\vspace{-0.5cm}
\end{figure*}

\textbf{Joint Torque Prediction Task}. We benchmark the representational capability of the latent state posterior, $\vec z_t^+$, of ac-RKN in accurately modelling the inverse dynamics of these robots in comparison to deterministic models like LSTMs and FFNN. It is clear from \ref{fig:frnaka_inv_torque} and \ref{fig:barett_inv} that ac-RKN learns highly accurate models of these in contrast with other data-driven methods. This highly precise modelling is often a requirement for high fidelity and compliant robotic control.   
% \begin{wrapfigure}[12]{r}{.27\linewidth}
% %      \hspace*{0.75cm} 
% %      \includegraphics[scale=0.5]{Figures/RKN-SchematicArrow.pdf}
%  %     \caption{Schematic Diagram Of Action Conditional Recurrent Kalman Network. Action conditioning takes place via additive interactions between the latent state $\vec{z}_{t}$ and action $a_t$ at the predict stage in the RKN Cell.}

% %     \hspace*{0.2cm} 
%   %   \includegraphics[scale=0.4]{Figures/ForwardArrow.pdf}
% %   \begin{subfigure}[b]{0.45\textwidth}
% %     \centering
% %      \resizebox{\textwidth}{!}{\tikzACRKN}
% %      \end{subfigure}
% % \qquad     
%     \begin{subfigure}[b]{0.27\textwidth}
%     \centering
%     \includegraphics[width=\linewidth]{Figures/rubber.pdf}
%      \end{subfigure}

% % \begin{subfigure}[b]{0.5\textwidth}
% %      \centering
% %      \resizebox{\textwidth}{!}{\tikzACRKN}
% %          \caption{}
% %      \end{subfigure}
% \caption{ } 
%  \label{fig:rubberband}
%  \end{wrapfigure} 
 
\textbf{Impact of Action Feedback}. We also perform an ablation study with and without the action feedback for the prediction step in the latent dynamics. As seen in the Figure \ref{fig:franka_inter}, the action feedback always results in better representational capability as this helps the implicit forward model in making better predictions by taking into account its causal effect on the state transitions.

\textbf{Comparison to Analytical Models}. Finally, we make a comparison with the analytical model of the Panda robot for inverse dynamics. Please refer to Appendix C for more details on the analytical model. As evident from Figure \ref{fig:inv_analytical} analytical models gave predictions with much lesser accuracy in comparison to ac-RKN, as it does not consider unmodelled effects such as joint and link flexibilities, backlash, stiction and actuator dynamics.  In such cases, the robot would continuously have to track its current position and compensate the errors with high-gain feedback control thus making it dangerous to interact with the real world and impossible to work in human-centred environments. % Nevertheless, analytical models often perform better on extrapolating to unseen tasks or work space configurations~\cite{hitzler2019learning} and hence as future work we intend to extend this framework to do online learning of filter parameters and benchmark its capability to extend its existing knowledge and adapt with new situations. 

\section{CONCLUSION}
We introduced a probabilistic recurrent neural network architecture, called action-conditional Recurrent Kalman Networks (ac-RKN), capable of action conditioning within the recurrent cell in a principled manner. This formulation allows us to learn highly accurate forward dynamics models of sophisticated robots and scenarios for which analytical models do not exist. We demonstrated the effectiveness of the proposed approach on robots with hydraulic, pneumatic and electric actuators. Besides, we leveraged the action-conditional recurrent cell to propose a regularized inverse dynamics model which significantly outperformed the current state of the art on two benchmarks. We believe the disentangled representation of the control signal in the ac-RKN recurrent cell has the potential to be exploited for several other robot learning problems. Also, since our architecture is domain-independent, we expect that they will generalize to several other robot dynamics learning tasks. As future work, we will employ these models for real-time control on some of the robotic systems, and we will learn models that predict future reward in addition to future states or observations and evaluate the performance of our approach in model-based reinforcement learning. Additionally, we will explore the incremental learning of Kalman Filter parameters to adapt to changing workspace configurations and tasks.


%%%%%%%%%%%%%%%%%%%%%%%%%%%%%%%%%%%%%%%%%%%%%%%%%%%%%%%%%%%%%%%%%%%%%%%%%%%%%%%%

%%%%%%%%%%%%%%%%%%%%%%%%%%%%%%%%%%%%%%%%%%%%%%%%%%%%%%%%%%%%%%%%%%%%%%%%%%%%%%%%

%%%%%%%%%%%%%%%%%%%%%%%%%%%%%%%%%%%%%%%%%%%%%%%%%%%%%%%%%%%%%%%%%%%%%%%%%%%%%%%%

%% Use plainnat to work nicely with natbib. 

%===============================================================================

% The maximum paper length is 8 pages excluding references and acknowledgements, and 10 pages including references and acknowledgements

\clearpage
% The acknowledgments are automatically included only in the final version of the paper.
\acknowledgments{This work was supported by the EPSRC UK (project   NCNR,   National Centre for Nuclear Robotics, EP/R02572X/1). We thank Aravinda Ramakrishnan Srinivasan for his support with the robots. }

%===============================================================================

\bibliography{example}  % .bib
\newpage

\title{Appendix}
\maketitle

% The \author macro works with any number of authors. There are two
% commands used to separate the names and addresses of multiple
% authors: \And and \AND.
%
% Using \And between authors leaves it to LaTeX to determine where to
% break the lines. Using \AND forces a line break at that point. So,
% if LaTeX puts 3 of 4 authors names on the first line, and the last
% on the second line, try using \AND instead of \And before the third
% author name.

% NOTE: authors will be visible only in the camera-ready (ie, when using the option 'final'). 
% 	For the initial submission the authors will be anonymized.

%%%%%%%%%%%%%%%%%%%%%%%%%%%%%%%%%%%%%%%%%%%%%%%%%%%%%%%%%%%%%%%%%%%%%%%%%%%%%%%%

%%%%%%%%%%%%%%%%%%%%%%%%%%%%%%%%%%%%%%%%%%%%%%%%%%%%%%%%%%%%%%%%%%%%%%%%%%%%%%%%

\section*{Appendix A: RKN - Summary and Conceptual Components}
The Recurrent Kalman Network (RKN) \cite{becker2019recurrent} integrates uncertainty estimates into deep time-series modelling by incorporating Kalman Filters~\cite{kalman1960new} into deep recurrent models. While Kalman filtering in the original state space requires approximations due to the non-linear models, the RKN uses a learned high-dimensional latent state representation that allows for efficient inference using locally linear transition models and a factorized belief state representation. The RKN consists of the following conceptual components.

\subsection*{Observation and Latent State Representation} The RKN transforms the observations at each time step $\vec{o}_t$ to a high dimensional space using an encoder network which emits high dimensional latent features $\vec{w}_t$ and an estimate of the uncertainty in those features via a variance vector $\vec{\sigma}_t^\mathrm{obs}$. %This idea is related to kernel methods \cite{gebhardt2017kernel} which use high-dimensional feature spaces to approximate nonlinear functions with linear models. However, in difference to kernel feature spaces, the RKN learns this feature space in an end-to-end manner using the deep encoder.\\
%\subsection{Latent State and Factorized Covariance Represenation} 

The probabilistic recurrent module uses a latent state vector $\vec{z}_t$ and corresponding covariance $\vec{\Sigma}_t$ whose transitions are governed by the Kalman Filter in the RKN memory cell. The latent state vector $\vec{z}_t$ has been designed to contain two conceptual parts, a vector $\vec{p}_t$ for holding information that can directly be extracted from the observations and a vector $\vec{m}_t$ to store information inferred over time, e.g., velocities. The former is referred to as the observation or upper part and the latter as the memory or lower part of the latent state by the authors \cite{becker2019recurrent}. For an ordinary dynamical system and images as observations the former may correspond to positions while the latter corresponds to velocities. The corresponding posterior and prior covariance matrices $\vec{\Sigma}_t^+$ and $\vec{\Sigma}_t^-$ have a chosen factorized representation to yield simple Kalman updates, i.e.,

\begin{equation*}
\vec{\Sigma}_t = 
\begin{bmatrix}
\vec{\Sigma}_t^\mathrm{u} & \vec{\Sigma}_t^\mathrm{s} \\
\vec{\Sigma}_t^\mathrm{s} & \vec{\Sigma}_t^\mathrm{l}  \\
\end{bmatrix},
\end{equation*}
where each of $\vec{\Sigma}_t^\mathrm{u}, \vec{\Sigma}_t^\mathrm{s}, \vec{\Sigma}_t^\mathrm{l} \in \mathbb{R}^{m \times m}$ is a diagonal matrix. The vectors  $\vec{\sigma}_t^\mathrm{u}, \vec{\sigma}_t^\mathrm{l}$ and $\vec{\sigma}_t^\mathrm{s}$ denote the vectors containing the diagonal values of those matrices. This structure with $\vec{\Sigma}_t^\mathrm{s}$ ensures that the correlation between the memory and the observation parts are not neglected as opposed to the case of designing $\vec{\Sigma}_t$ as a diagonal covariance matrix. This representation was exploited to avoid the expensive and numerically problematic matrix inversions involved in the KF equations as shown below. \\

\subsection*{Locally Linear Transition Model} The state transitions in the predict stage of the Kalman filter is governed by a locally linear transition model. To obtain a locally linear transition model, the RKN learns $K$ constant transition matrices $\vec{A}^{(k)}$ and combines them using state dependent coefficients $\alpha^{(k)}(\vec{z_t})$, i.e.,
$ \vec{A}_t = \sum_{k=0}^K \alpha^{(k)}(\vec{z_t}) \vec{A}^{(k)}. $
A small neural network with softmax output is used to learn $\alpha^{(k)}$. Each $\vec{A}^{(k)}$ is designed to consist of four band matrices as in \cite{becker2019recurrent} in order to reduce the number parameters without affecting the performance.\\

\subsection*{Observation Model} The latent state space $\mathcal{Z} = \mathbb{R}^n$ of the RKN is related to the observation space $\mathcal{W}$ by the linear latent observation model $\vec{H} = \left[\begin{array}{cc} \vec{I}_m & \vec{0}_{m \times (n-m)} \end{array}\right]$, i.e., $ \vec{w} = \vec{H} \vec{z}$ with  $\vec{w} \in \mathcal{W}$ and $\vec{z} \in \mathcal{Z}$, where $\vec{I}_m$ denotes the $m \times m$ identity matrix and $\vec{0}_{m \times (n-m)}$ denotes a $m \times (n-m)$ matrix filled with zeros. Typically, $m$ is set to $n / 2$. This corresponds to the assumption that the first half of the state can be directly observed while the second half is unobserved and contains information inferred over time. \\ 

\subsection*{Kalman Update Step} The Kalman update involves computing the Kalman gain matrix $\vec{Q}_t$, which requires computationally expensive matrix inversions that are difficult to backpropagate, at least for high dimensional latent state representations. However, the choice of a locally linear transition model, the factorized covariance $\vec{\Sigma}_t$, and the special observation model simplify the Kalman update to scalar operations as shown below. As the network is free to choose its own state representation, it finds a representation where such assumptions works well in practice \cite{becker2019recurrent}.  

Similar to the state, the Kalman gain matrix $\vec{Q}_t$ is split into an upper $\vec{Q}_t^\textrm{u}$ and a lower part $\vec{Q}_t^\textrm{l}$. Both $\vec{Q}_t^\textrm{u}$ and $\vec{Q}_t^\textrm{l}$ are squared matrices. Due to the simple latent observation model $\vec{H} = \left[\begin{array}{cc} \vec{I}_m & \vec{0}_{m \times (n-m)} \end{array}\right]$ and the factorized covariances, all off-diagonal entries of $\vec{Q}_t^\textrm{u}$ and $\vec{Q}_t^\textrm{l}$ are zero and one can again work with vectors representing the diagonals, i.e., $\vec{q}_t^\textrm{u}$ and $\vec{q}_t^\textrm{l}$. Those are obtained by
\begin{align*}\vec{q}_t^\mathrm{u} &= \vec{\sigma}_t^{\mathrm{u},-} \oslash \left( \vec{\sigma}_t^{\mathrm{u},-} + \vec{\sigma}_t^\mathrm{obs}  \right) \\
\vec{q}_t^\mathrm{l} &= \vec{\sigma}_t^{\mathrm{s},-} \oslash \left( \vec{\sigma}_t^{\mathrm{u},-} + \vec{\sigma}_t^\mathrm{obs} \right),
\end{align*}
where $\oslash$ denotes an elementwise vector division. The update equation for the mean therefore simplifies to 
\begin{align*}
\vec{z}_t^+ = \vec{z}_t^- +
\left[\begin{array}{c} \vec{q}^\mathrm{u}_t \\
\vec{q}^\mathrm{l}_t \end{array}\right]
\odot
\left[\begin{array}{c}\vec{w}_t - \vec{z}^{\mathrm{u},-}_t \\
\vec{w}_t - \vec{z}^{\mathrm{u},-}_t  \end{array}\right],
\end{align*}
where $\odot$ denotes the elementwise vector product. The update equations for the individual parts of covariance are given by
\begin{align*}
\vec{\sigma}^{\mathrm{u},+}_t &= \left( \vec{1}_m - \vec{q}^\mathrm{u}_t \right) \odot \vec{\sigma}^{\mathrm{u},-}_t,  \\
\vec{\sigma}^{\mathrm{s},+}_t &= \left( \vec{1}_m - \vec{q}^\mathrm{u}_t \right) \odot \vec{\sigma}^{\mathrm{s},-}_t, \\
\vec{\sigma}^{\mathrm{l},+}_t &= \vec{\sigma}^{\mathrm{l}, -}_t - \vec{q}^\mathrm{l}_t \odot \vec{\sigma}^{\mathrm{s},-}_t,
\end{align*}
where $\vec{1}_m$ denotes the $m$ dimensional vector consisting of ones.
   
\section*{Appendix B: Robots and Data}

The experiments are performed on data from four different robots. The details of robots, data and data preprocessing is explained below:
\begin{figure*}[h!t!]
\centering
\includegraphics[width=4.1cm,height=3.5cm]{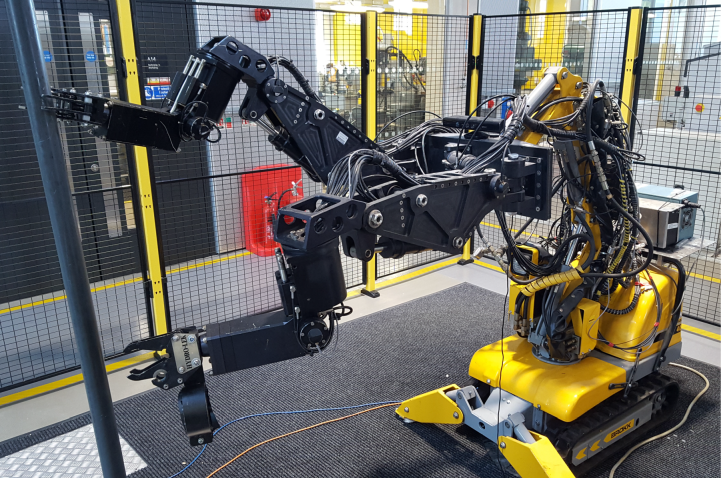}
\includegraphics[width=4.1cm,height=3.5cm]{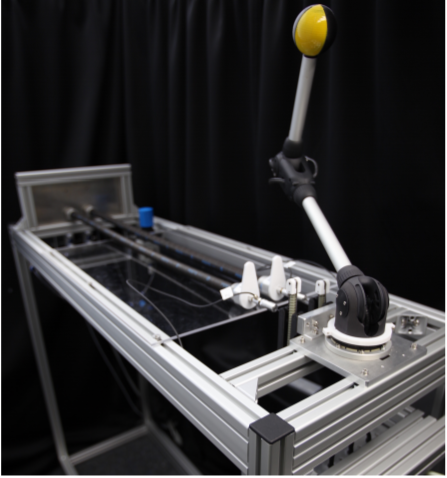}
\includegraphics[width=4.1cm,height=3.5cm]{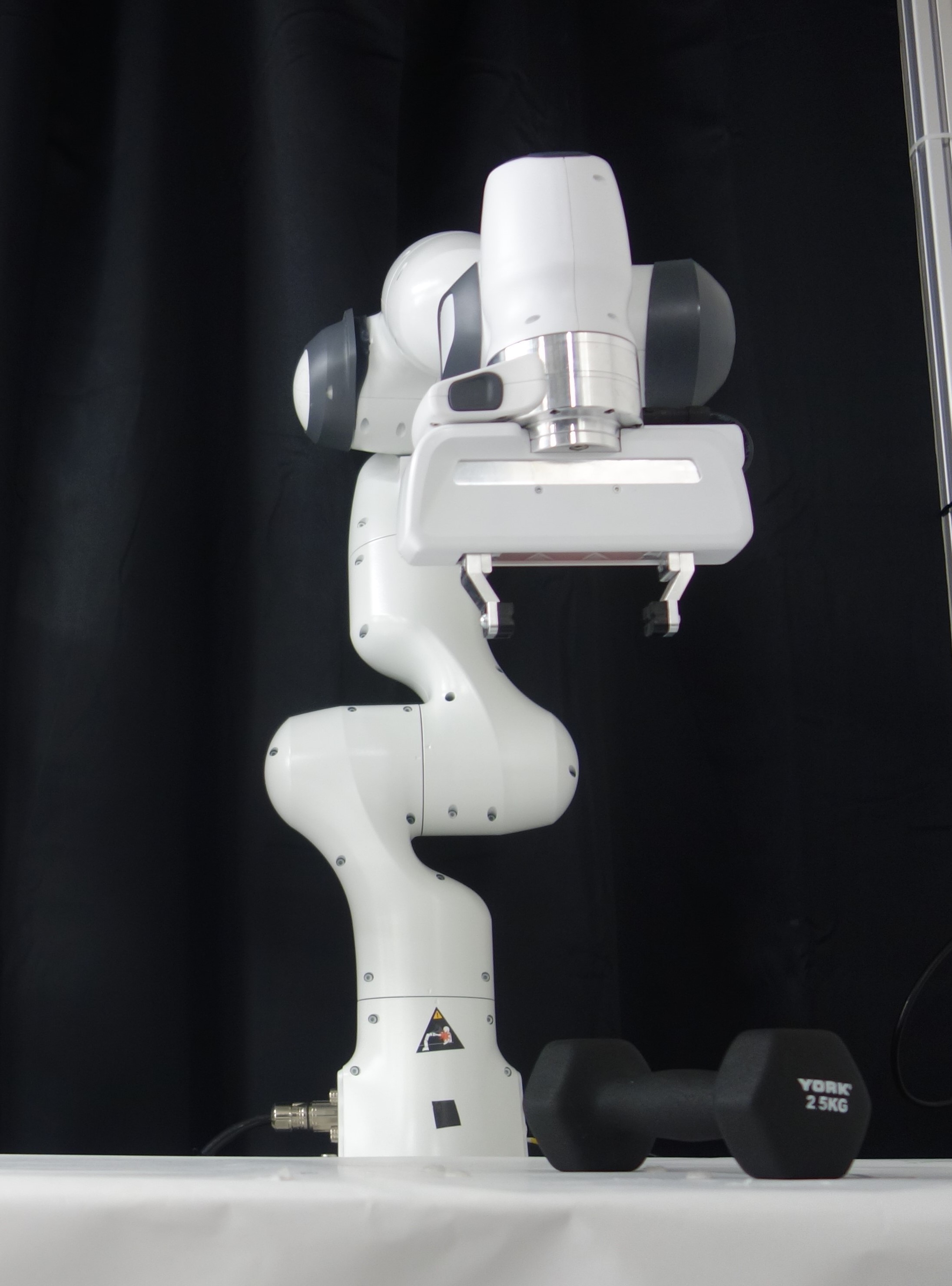}
\vspace{0.1cm}  
\caption{The experiments are performed on data from robots with different actuator dynamics. From left to right these include: Hydraulically actuated BROKK-40 \cite{taylor2013state}, Pneumatically actuated artificial muscles \cite{buchler2016lightweight}, Franka Emika Panda Robotic Arm.}
\label{fig:robots}
\vspace{-0.5cm}
\end{figure*}
\subsection*{Hydraulic Brokk 40 Robot Arm}

\textbf{Observation and Data Set}: The data was obtained from a HYDROLEK–7W 6 degree–of–freedom manipulator with a continuous (360 degree) jaw rotation mechanism. We actuate the joints via hydraulic pistons, which are powered via an auxiliary output from the hydraulic pump. Thus learning the forward model is difficult due to inherent hysteresis associated with hydraulic control. For this robot, only one joint is moved at a time, so we have independent time series per joint. The joint data consists of measured joint positions and the input current to the controller of the joint sampled at 100Hz. \\

\textbf{Training Procedure}: During training, we work with sequences of length 500. For the first 300 time steps those sequences consist of the full observation, i.e., the joint position and current. We give only the current signals in the remaining 200 time steps. The models have to impute the missing joint positions in an uninformed fashion, i.e., we only indicate the absence of a position by unrealistically high values.

\subsection*{Musculoskeletal Robot Arm}
\label{sec:PAMdesc}

\textbf{Observation and Data Set:}  For this soft robot we have 4 dimensional observation inputs(joint angles) and 8 dimensional action inputs(pressures). We collected the data of a four DoF robot actuated by Pneumatic Artificial Muscles
(PAMs). The robot arm has eight PAMs in total with each DoF actuated by an antagonistic pair. The robot arm reaches high joint angle accelerations of up to 28, 000deg/s2 while avoiding dangerous joint limits thanks to the antagonistic actuation and limits on the air pressure ranges. The data
consists of trajectories collected while training with a model-free reinforcement learning algorithm to hit balls while playing table tennis. We sampled the data at 100Hz. The hysteresis associated with the pneumatic actuators used in this robot is challenging to model and is relevant to the soft robotics in general.\\

\textbf{Training Procedure}: During training, we randomly removed three-quarters of the states from the sequences and tasked the models with imputing those missing states, only based on the knowledge of available actions/control commands, i.e., we train the models to perform action conditional future predictions to impute missing states. The imputation employs the model for multi-step ahead predictions in a convenient way. One could instead go for a dedicated loss function as in approaches like \cite{finn2016unsupervised}, \cite{oh2015action} for long term predictions.

\subsubsection*{Franka Emika Panda Robot Arm}

\textbf{Observation and Data Set:} We collected the data from a 7 DoF Franka Emika Panda manipulator during free motion. We chose this task since the robot exhibits different dynamics behaviour due to electric actuators and high frequencies(1kHz). The raw joint positions, velocities and torques were recorded using Franka Interfaces while the joint accelerations were computed by finite differences on filtered velocity data (obtained using a zero-phase 8th-order digital Butterworth filter with a cut-off frequency of 5Hz). The observations for the forward model consist of the seven joint angles in radians, and the corresponding actions were joint Torques in Nm. While the inverse model use both joint angles and velocities as observations. The data was divided into train and test sets in the ratio 4:1. We divide the data into sequences of length 300 while training the recurrent models for forward dynamics and use sequences of length 50 for inverse dynamics.   \\  

\textbf{Training Procedure Forward Dynamics}: Similar to the multi-step ahead training procedure in \ref{sec:PAMdesc}, during training we randomly removed three-quarters of the observations(joint angles) from the sequences and tasked the models with imputing those missing observations, only based on the knowledge of available actions/control commands.

\textbf{Training Procedure Inverse Dynamics}: The recurrent models (LSTM, ac-RKN) uses a similar architecture, as shown in Figure 3 of the main paper, except for the recurrent module. The hyperparameters including learning rate, latent state and observation dimensions, learning rate, control model architecture, action decoder architecture and regularization parameter for the joint forward-inverse dynamics loss function are searched via GpyOpt\cite{gpyopt2016} and is mentioned in Appendix D. The observation encoder and decoder architecture is chosen to be of the same size across the models being compared. For all models, we use the joint positions and velocities as the observation input and differences to the next state as desired observation. The FFNN gets the current observation and desired observation as input and is tasked to predict the joint Torques directly(unlike differences in recurrent models) as in previous regression approaches\cite{nguyen2010using}. 
\label{sec:invtraining}

\subsubsection*{Barrett WAM Robot Arm}

\textbf{Observation and Data Set:} The Barett task is based on a publicly available dataset comprising joint positions, velocities, acceleration and torques of a seven degrees-of-freedom real Barett WAM robot. The original training dataset (12, 000 data points) is split into sequences of length 98. Twenty-four out of the total 119 episodes are utilized for testing, whereas the other 95 are used for training. The direct cable drives which drive this robot produce high torques, generating fast and dexterous movements but yield complex dynamics.  Rigid-body dynamics cannot be model this complex dynamics due to the variable stiffness and lengths of the cables. \\  

\textbf{Training Procedure Inverse Dynamics}: The training procedure is repeated as in \ref{sec:invtraining}

\section*{Appendix C: Details Of Rigid Body Dynamics Model}
The analytical model for Franka Emika Panda is a rigid-body dynamics model that was identified in its so-called base parameters~\cite{Khalil1991}. Due to the friction compensation in the joints, we observed that the viscous friction is negligible, whereas the observed Coulomb friction is very small yet included in our parameterization. Which results is a model with 50 parameters. 
The base parameterization is computed based on provided kinematic properties of the robotic arm and provides a linear relation between the base parameters and joint torques for a given set of joint positions, velocities and accelerations. 

Due to this linearity, the regression problem can be solved using a linear least-squares method, although additional linear matrix inequality constraints must be fullfilled to ensure that the resulting parameters are physically realizable \cite{Sousa2019}.
In order to perform forward simulation of the robot dynamics, we numerically solve an initial value problem for the implicit set of differential equations defined by base parameterization of the rigid-body model.
Note that, the model does not parameterize actuator dynamics, nor does it model joint flexibilities, link flexibilities, or stiction. The focus here is to provide a reference baseline to show which effects the acRKN captures in comparison to a text-book robot model.

\section*{Appendix D: Hyperparameters}

\subsection*{Pneumatic Musculoskeltal Robot Arm}
\begin{longtable}{|l|l|l|l|}
\caption{Forward Dynamics Hyperparameters For Pneumatic Musculoskeltal Robot.} \label{tab:long} \\

\hline \multicolumn{1}{|c|}{\textbf{Hyperparameter}} & \multicolumn{1}{c|}{\textbf{ac-RKN}} & \multicolumn{1}{c|}{\textbf{RKN}} & \multicolumn{1}{c|}{\textbf{LSTM}}  \\ \hline 
\endfirsthead

\multicolumn{3}{c}%
{{\bfseries \tablename\ \thetable{} -- continued from previous page}} \\
\hline \multicolumn{1}{|c|}{\textbf{First column}} & \multicolumn{1}{c|}{\textbf{Second column}} & \multicolumn{1}{c|}{\textbf{Third column}} & \multicolumn{1}{c|}{\textbf{fOURTH column}}\\ \hline 
\endhead

\hline \multicolumn{3}{|r|}{{Continued on next page}} \\ \hline
\endfoot

\hline
\endlastfoot

Learning Rate & 3.1e-3 & 1.9e-3 &6.6e-3 \\ \hline
Latent Observation Dimension & 60 & 60 & 60  \\ \hline
Latent State Dimension & 120 & 120 & 120  \\
\end{longtable}

\underline{Encoder} (ac-RKN,RKN,LSTM): 1 fully connected + linear output  (elu + 1) 
\begin{itemize}
	\item Fully Connected 1: 120, ReLU
\end{itemize} 
\underline{Observation Decoder} (ac-RKN,RKN,LSTM): 1 fully connected + linear output:
\begin{itemize}
	\item Fully Connected 1: 120, ReLU
\end{itemize} 
 
\underline{Transition Model} (ac-RKN,RKN):
bandwidth: 3, number of basis: 15
\begin{itemize}
    \item $\alpha(\vec{z}_t)$: No hidden layers - softmax output
\end{itemize}
\underline{Control Model} (ac-RKN):
3 fully connected + linear output
\begin{itemize}
	\item Fully Connected 1: 120, ReLU
	\item Fully Connected 2: 120, ReLU
	\item Fully Connected 3: 120, ReLU
\end{itemize} 

\textbf{Architecture For FFNN Baseline}
2 fully connected + linear output
\begin{itemize}
	\item Fully Connected 1: 6000, ReLU
	\item Fully Connected 2: 3000, ReLU
\end{itemize}

Dropout Regularization - 0.512\\
Learning Rate - 1.39e-2\\
Optimizer Used: Adam Optimizer

\subsection*{Hydraulic Brokk-40 Robot Arm}
\begin{longtable}{|l|l|l|l|l|}
\caption{Forward Dynamics Hyperparameters For Pneumatic Musculoskeltal Robot.} \label{tab:long} \\

\hline \multicolumn{1}{|c|}{\textbf{Hyperparameter}} & \multicolumn{1}{c|}{\textbf{ac-RKN}} & \multicolumn{1}{c|}{\textbf{RKN}} & \multicolumn{1}{c|}{\textbf{LSTM}} &
\multicolumn{1}{c|}{\textbf{GRU}}  \\ \hline 
\endfirsthead

\multicolumn{3}{c}%
{{\bfseries \tablename\ \thetable{} -- continued from previous page}} \\
\hline \multicolumn{1}{|c|}{\textbf{First column}} & \multicolumn{1}{c|}{\textbf{Second column}} & \multicolumn{1}{c|}{\textbf{Third column}} & \multicolumn{1}{c|}{\textbf{fOURTH column}}\\ \hline 
\endhead

\hline \multicolumn{3}{|r|}{{Continued on next page}} \\ \hline
\endfoot

\hline
\endlastfoot

Learning Rate & 5e-4 & 5e-4 &9.1e-4 &2.1e-3 \\ \hline
Latent Observation Dimension & 30 & 30 & 30 & 30  \\ \hline
Latent State Dimension & 60 & 60 & 60 & 60  \\
\end{longtable}

\underline{Encoder} (ac-RKN,RKN,LSTM,GRU): 1 fully connected + linear output  (elu + 1) 
\begin{itemize}
	\item Fully Connected 1: 30, ReLU
\end{itemize} 
\underline{Observation Decoder} (ac-RKN,RKN,LSTM,GRU): 1 fully connected + linear output:
\begin{itemize}
	\item Fully Connected 1: 30, ReLU
\end{itemize} 
 
\underline{Transition Model} (ac-RKN,RKN):
bandwidth: 3, number of basis: 32
\begin{itemize}
    \item $\alpha(\vec{z}_t)$: No hidden layers - softmax output
\end{itemize}
\underline{Control Model} (ac-RKN):
1 fully connected + linear output
\begin{itemize}
	\item Fully Connected 1: 120, ReLU
\end{itemize} 

\subsection*{Franka Emika Panda - Forward Dynamics Learning}
\begin{longtable}{|l|l|l|l|l|}
\caption{Forward Dynamics Learning Hyperparameters For Panda.} \label{tab:long} \\

\hline \multicolumn{1}{|c|}{\textbf{Hyperparameter}} & \multicolumn{1}{c|}{\textbf{ac-RKN}} & \multicolumn{1}{c|}{\textbf{RKN}} & \multicolumn{1}{c|}{\textbf{LSTM}} & \multicolumn{1}{c|}{\textbf{GRU}} \\ \hline 
\endfirsthead

\multicolumn{3}{c}%
{{\bfseries \tablename\ \thetable{} -- continued from previous page}} \\
\hline \multicolumn{1}{|c|}{\textbf{First column}} & \multicolumn{1}{c|}{\textbf{Second column}} & \multicolumn{1}{c|}{\textbf{Third column}} & \multicolumn{1}{c|}{\textbf{fOURTH column}}\\ \hline 
\endhead

\hline \multicolumn{3}{|r|}{{Continued on next page}} \\ \hline
\endfoot

\hline
\endlastfoot

Learning Rate & 3.1e-3 & 1.7e-3 &6.6e-3 &8.72e-3\\ \hline
Latent Observation Dimension & 45 & 30 & 30 & 45 \\ \hline
Latent State Dimension & 90 & 60 & 60 & 90 \\
\end{longtable}

\underline{Encoder} (ac-RKN,RKN,LSTM,GRU): 1 fully connected + linear output  (elu + 1) 
\begin{itemize}
	\item Fully Connected 1: 120, ReLU
\end{itemize} 
\underline{Observation Decoder} (ac-RKN,RKN,LSTM,GRU): 1 fully connected + linear output:
\begin{itemize}
	\item Fully Connected 1: 240, ReLU
\end{itemize} 
 
\underline{Transition Model} (ac-RKN,RKN):
bandwidth: 3, number of basis: 15
\begin{itemize}
    \item $\alpha(\vec{z}_t)$: No hidden layers - softmax output
\end{itemize}
\underline{Control Model} (ac-RKN):
3 fully connected + linear output
\begin{itemize}
	\item Fully Connected 1: 30, ReLU
	\item Fully Connected 2: 30, ReLU
	\item Fully Connected 3: 30, ReLU
\end{itemize} 

\textbf{Architecture For FFNN Baseline - Forward Dynamics}
3 fully connected + linear output
\begin{itemize}
	\item Fully Connected 1: 1000, ReLU
	\item Fully Connected 2: 1000, ReLU
	\item Fully Connected 3: 1000, ReLU
\end{itemize}

Dropout Regularization - 0.1147\\
Learning Rate - 8.39e-3\\
Optimizer Used: SGD Optimizer

\subsection*{Franka Emika Panda - Inverse Dynamics Learning}
\begin{longtable}{|l|l|l|l|l|}
\caption{Inverse Dynamics Learning Hyperparameters For Panda.} \label{tab:long} \\

\hline \multicolumn{1}{|c|}{\textbf{Hyperparameter}} & \multicolumn{1}{c|}{\textbf{ac-RKN}} & \multicolumn{1}{c|}{\textbf{RKN (No Action Feedback)}} & \multicolumn{1}{c|}{\textbf{LSTM}} \\ \hline 
\endfirsthead

\multicolumn{3}{c}%
{{\bfseries \tablename\ \thetable{} -- continued from previous page}} \\
\hline \multicolumn{1}{|c|}{\textbf{First column}} & \multicolumn{1}{c|}{\textbf{Second column}} & \multicolumn{1}{c|}{\textbf{Third column}} & \multicolumn{1}{c|}{\textbf{fOURTH column}}\\ \hline 
\endhead

\hline \multicolumn{3}{|r|}{{Continued on next page}} \\ \hline
\endfoot

\hline
\endlastfoot

Learning Rate & 7.62e-3 & 3.5e-3 &9.89e-3 \\ \hline
Latent Observation Dimension & 15 & 30 & 30\\ \hline
Latent State Dimension & 30 & 60 & 60 \\ \hline
Regularization Factor ($\lambda$) &0.158 &0.179 &0.196\\
\end{longtable}

\underline{Encoder} (ac-RKN,RKN,LSTM): 1 fully connected + linear output (elu + 1) 
\begin{itemize}
	\item Fully Connected 1: 120, ReLU
\end{itemize} 
\underline{Observation Decoder} (ac-RKN,RKN,LSTM): 1 fully connected + linear output:
\begin{itemize}
	\item Fully Connected 1: 240, ReLU
\end{itemize} 

\underline{Action Decoder} (ac-RKN,RKN,LSTM): 1 fully connected + linear output:
\begin{itemize}
	\item Fully Connected 1: 512, ReLU
\end{itemize} 
 
\underline{Transition Model} (ac-RKN,RKN):
bandwidth: 3, number of basis: 15
\begin{itemize}
    \item $\alpha(\vec{z}_t)$: No hidden layers - softmax output  
\end{itemize}

\underline{Control Model} (ac-RKN):
1 fully connected + linear output
\begin{itemize}
	\item Fully Connected 1: 45, ReLU
\end{itemize} 

\textbf{Architecture For FFNN Baseline - Inverse Dynamics}
3 fully connected + linear output
\begin{itemize}
	\item Fully Connected 1: 500, ReLU
	\item Fully Connected 2: 500, ReLU
	\item Fully Connected 3: 500, ReLU
\end{itemize}

Dropout Regularization - 0.563 \\
Learning Rate - 1.39e-2 \\
Optimizer Used: SGD Optimizer

\subsection*{Barett WAM - Inverse Dynamics Learning}
\begin{longtable}{|l|l|l|l|l|}
\caption{Inverse Dynamics Learning Hyperparameters Barett WAM.} \label{tab:long} \\

\hline \multicolumn{1}{|c|}{\textbf{Hyperparameter}} & \multicolumn{1}{c|}{\textbf{ac-RKN}} & \multicolumn{1}{c|}{\textbf{RKN (No Action Feedback)}} & \multicolumn{1}{c|}{\textbf{LSTM}}  \\ \hline 
\endfirsthead

\multicolumn{3}{c}%
{{\bfseries \tablename\ \thetable{} -- continued from previous page}} \\
\hline \multicolumn{1}{|c|}{\textbf{First column}} & \multicolumn{1}{c|}{\textbf{Second column}} & \multicolumn{1}{c|}{\textbf{Third column}} & \multicolumn{1}{c|}{\textbf{fOURTH column}}\\ \hline 
\endhead

\hline \multicolumn{3}{|r|}{{Continued on next page}} \\ \hline
\endfoot

\hline
\endlastfoot

Learning Rate & 7.7e-3 & 1.7e-3 &9.33e-3 \\ \hline
Latent Observation Dimension & 15 & 30 & 45  \\ \hline
Latent State Dimension & 30 & 60 & 90 \\ \hline
Regularization Factor ($\lambda$) &0.176 &0 &3.42e-3\\
\end{longtable}

\underline{Encoder} (ac-RKN,RKN,LSTM): 1 fully connected + linear output  (elu + 1) 
\begin{itemize}
	\item Fully Connected 1: 120, ReLU
\end{itemize} 
\underline{Observation Decoder} (ac-RKN,RKN,LSTM): 1 fully connected + linear output:
\begin{itemize}
	\item Fully Connected 1: 240, ReLU
\end{itemize} 

\underline{Action Decoder} (ac-RKN): 2 fully connected + linear output:
\begin{itemize}
	\item Fully Connected 1: 256, ReLU
	\item Fully Connected 1: 256, ReLU
\end{itemize} 

\underline{Action Decoder} (RKN,LSTM): 1 fully connected + linear output:
\begin{itemize}
	\item Fully Connected 1: 512, ReLU
\end{itemize} 
 
\underline{Transition Model} (ac-RKN,RKN):
bandwidth: 3, number of basis: 15
\begin{itemize}
    \item $\alpha(\vec{z}_t)$: No hidden layers - softmax output
\end{itemize}

\underline{Control Model} (ac-RKN):
1 fully connected + linear output
\begin{itemize}
	\item Fully Connected 1: 45, ReLU
\end{itemize} 

\textbf{Architecture For FFNN Baseline}
3 fully connected + linear output
\begin{itemize}
	\item Fully Connected 1: 500, ReLU
	\item Fully Connected 2: 500, ReLU
	\item Fully Connected 3: 500, ReLU
\end{itemize}

Dropout Regularization - 0.563\\
Learning Rate - 1e-5\\
Optimizer Used: SGD Optimizer

\end{document}